%% file: main.tex
\documentclass[10pt,twocolumn,letterpaper]{article}

\usepackage{cvpr}              

\usepackage{graphicx}
\usepackage{amsmath}
\usepackage{amssymb}
\usepackage{booktabs}
\usepackage{multirow}
\newcommand{\nn}{MCTformer}

\usepackage{adjustbox}

\usepackage[pagebackref,breaklinks,colorlinks]{hyperref}

\usepackage[capitalize]{cleveref}
\crefname{section}{Sec.}{Secs.}
\Crefname{section}{Section}{Sections}
\Crefname{table}{Table}{Tables}
\crefname{table}{Tab.}{Tabs.}

\def\cvprPaperID{5253} 
\def\confName{CVPR}
\def\confYear{2022}

\input{sec/preamble}
\begin{document}

\input{sec/0_metadata}
\maketitle
\input{sec/0_abstract}
\input{sec/1_introduction}
\input{sec/2_related}

\input{sec/3_method}
\input{sec/4_results}
\input{sec/5_conclusions}

{
    \small
    \bibliographystyle{ieee_fullname}
    \bibliography{macros,main}
}
\input{sec/X_supplementary}


\end{document}

%% file: sec/preamble.tex
\usepackage{overpic}
\usepackage{enumitem} 
\usepackage{overpic} 
\usepackage{color}

\definecolor{turquoise}{cmyk}{0.65,0,0.1,0.3}
\definecolor{purple}{rgb}{0.65,0,0.65}
\definecolor{dark_green}{rgb}{0, 0.5, 0}
\definecolor{orange}{rgb}{0.8, 0.6, 0.2}
\definecolor{red}{rgb}{0.8, 0.2, 0.2}
\definecolor{darkred}{rgb}{0.6, 0.1, 0.05}
\definecolor{blueish}{rgb}{0.0, 0.3, .6}
\definecolor{light_gray}{rgb}{0.7, 0.7, .7}
\definecolor{pink}{rgb}{1, 0, 1}
\definecolor{greyblue}{rgb}{0.25, 0.25, 1}

\newcommand{\Figure}[1]{Figure~\ref{fig:#1}}

\newcommand{\Table}[1]{Table~\ref{tab:#1}}
\newcommand{\eq}[1]{(\ref{eq:#1})}

\usepackage{blindtext}

\renewcommand{\paragraph}[1]{\vspace{1em}\noindent\textbf{#1}.}

%% file: sec/0_metadata.tex
\title{Multi-class Token Transformer for Weakly Supervised Semantic Segmentation}
\author{
      Lian Xu$^1$,
      Wanli Ouyang$^2$,
      Mohammed Bennamoun$^1$,
      Farid Boussaid$^1$,
      and
      Dan Xu$^3$%
    \\
    $^1$The University of Western Australia \quad $^2$The University of Sydney \\
\quad $^3$Hong Kong University of Science and Technology%
    \\
    \small{
    \tt{\{lian.xu,mohammed.bennamoun,farid.boussaid\}@uwa.edu.au}}, \\ 
    \small{\tt{wanli.ouyang@sydney.edu.au},  \tt{danxu@cse.ust.hk}}
}


%% file: sec/0_abstract.tex
\begin{abstract}

This paper proposes a new transformer-based framework to learn class-specific object localization maps as pseudo labels for weakly supervised semantic segmentation (WSSS). 
Inspired by the fact that the attended regions of the one-class token in the standard vision transformer can be leveraged to form a class-agnostic localization map, we investigate if the transformer model can also effectively capture class-specific attention for more discriminative object localization by learning multiple class tokens within the transformer. To this end, we propose a Multi-class Token Transformer, termed as \textbf{\nn}, which uses multiple class tokens to learn interactions between the class tokens and the patch tokens.
The proposed \nn~can successfully produce class-discriminative object localization maps from class-to-patch attentions corresponding to different class tokens. 
We also propose to use a patch-level pairwise affinity, which is extracted from the patch-to-patch transformer attention, to further refine the localization maps.
Moreover, the proposed framework is shown to fully complement the Class Activation Mapping (CAM) method, leading to remarkably superior WSSS results on the PASCAL VOC and MS COCO datasets.
These results underline the importance of the class token for WSSS.
~\footnote{\url{https://github.com/xulianuwa/MCTformer}}

\end{abstract}

%% file: sec/1_introduction.tex
\section{Introduction}
\label{sec:intro}
\input{fig/teaser}
\par Weakly supervised semantic segmentation (WSSS) aims to alleviate the reliance on pixel-level ground-truth labels by using weak supervision. A critical step for this task is to generate high-quality pseudo segmentation ground-truth labels by using weak labels. Image-level labels can provide simple weak labels which only indicate the presence or absence of certain classes without any ground-truth localization information. 
Previous WSSS methods generally rely on Class Activation Mapping (CAM)~\cite{zhou2016learning} to extract object localization maps from Convolutional Neural Networks (CNNs). 
Despite using complex CAM expansion strategies or multiple training steps, existing methods still exhibit limited performance in terms of both completeness of the localized objects and accuracy.
Vision Transformer (ViT)~\cite{dosovitskiy2020image}, as the first transformer model specifically designed for computer vision, has recently achieved performance breakthroughs on multiple vision tasks~\cite{khan2021transformers}. Particularly, ViT has achieved state-of-the-art performance for large-scale image recognition, thanks to its strong capability to model long-range contexts. ViT splits the input image into non-overlapping patches and transforms them into a sequence of vectors.
ViT also uses \emph{one} extra class token to aggregate information from the entire sequence of the patch tokens. 
Although the class token has been removed in a number of recent transformer methods~\cite{pan2021scalable,chu2021conditional, chu2021twins}, this work will underline its importance for weakly supervised semantic segmentation.

A recent work, DINO~\cite{caron2021emerging}, revealed that there was explicit information about the semantic segmentation of an image in self-supervised ViT features. More specifically, it was observed that a semantic scene layout can be discovered from the attention maps of the class token. These attention maps lead to promising results in the unsupervised segmentation task.  
Although it was demonstrated that different heads in the transformer attention can attend to different semantic regions of an image, it remains unclear how to associate a head to a correct semantic class. That is, these attention maps are still class-agnostic (see \Figure{teaser}).

\par It is challenging to exploit class-specific attention from transformers. We argue that existing transformer-based works have a common issue, \ie, using only \emph{one} class token, which makes the accurate localization of different objects on a single image challenging. There are two main reasons for this. \textbf{First}, a one-class-token design essentially 
inevitably captures context information from other object categories and the background. In other words, it naturally learns both class-specific and generic representations for different object classes as only one class token is considered, thus resulting in a rather non-discriminative and noisy object localization. \textbf{Second}, the model uses the only one-class token to learn interactions with patch tokens for 
a number of distinct object classes in a dataset. The model capacity is consequently not adequate enough to achieve 
the targeted discriminative localization performance.

To tackle these issues, a straightforward idea is to leverage multiple class tokens, which will be responsible for learning representations for different object classes.
To this end, we propose a Multi-class Token Transformer (\nn), 
in which \emph{multiple} class-specific tokens are employed to exploit class-specific transformer attention. Our goal of having class-specific tokens cannot be achieved by simply increasing the number of class tokens in ViT, because these class tokens still do not have specific meanings. To ensure that each class token can effectively learn high-level discriminative representations of a specific object class, we propose a class-aware training strategy for multiple class tokens. More specifically, we apply average pooling on the output class tokens from the transformer encoder along the embedding dimension, to generate class scores, which are directly supervised by the ground-truth class labels. This thus builds a one-to-one strong connection between each class token and the corresponding class label.
Through this design, one significant advantage is that the learned class-to-patch attention of different classes can be directly used as class-specific localization maps.  

It is worth noting that the learned patch-to-patch attention, as a byproduct of training without additional computation, can serve as a patch-level pairwise affinity. This can be used to further refine the class-specific transformer attention maps, dramatically improving the localization performance. 
Moreover, we also show that the proposed transformer framework fully complements the CAM method when applied on patch tokens (by simultaneously learning to classify with class-token and patch-token based representations).
This leads to high consistency between class tokens and patch tokens, thus considerably enhancing the discriminative ability of their derived object localization maps.

In summary, the main contribution is three-fold:
\begin{itemize}
    \item We propose to exploit class-specific transformer attentions for weakly supervised semantic segmentation. 
    \item We propose an effective transformer framework, which includes a novel multi-class token transformer (\nn) coupled with a class-aware training strategy, to learn class-specific localization maps from the class-to-patch attention of different class tokens.
    \item We propose to use the patch-to-patch transformer attentions as a patch-level pairwise affinity, which can significantly refine the class-specific transformer attentions. Furthermore, the proposed \nn~can fully complement the CAM mechanism, leading to high-quality object localization maps.
\end{itemize}
The proposed method can generate high-quality class-specific multi-label localization maps for WSSS, establishing new state-of-the-art results on PASCAL VOC (mIoU of 71.6\% on the test set) and MS COCO (mIoU of 42.0\%).

%% file: fig/teaser.tex
\begin{figure}[t]
\begin{center}
\includegraphics[width=\linewidth]{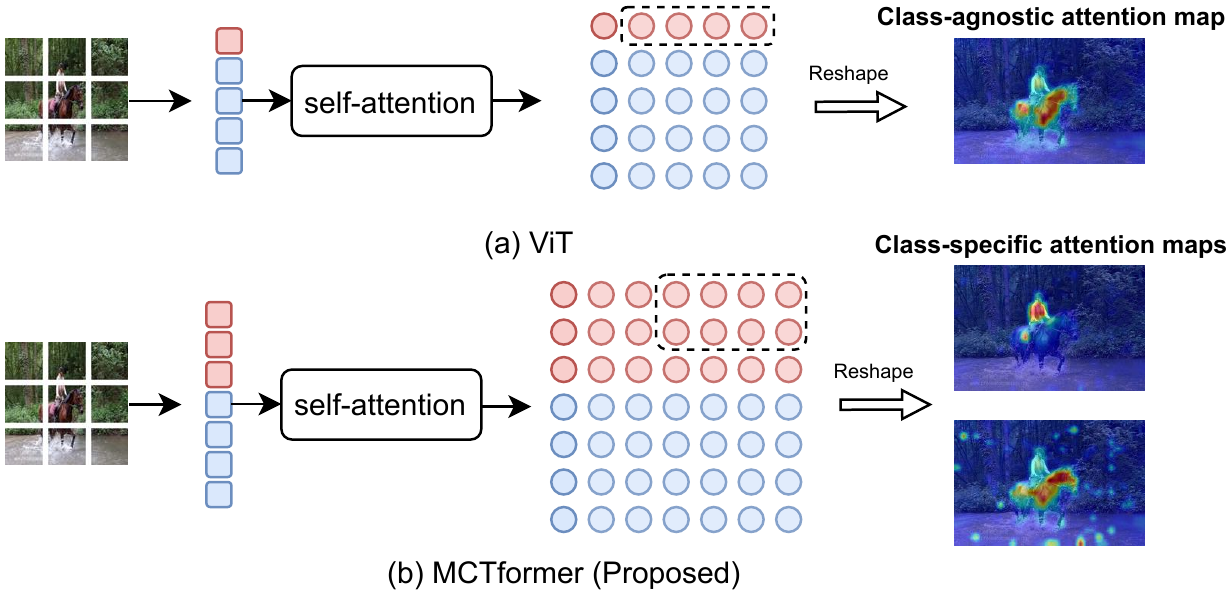}
\end{center}
\vspace{-15pt}
\caption{
%
(a) In previous vision transformers~\cite{dosovitskiy2020image}, only one class token (red square) is used to aggregate information from patch tokens (blue square). The learned patch attentions corresponding to the class token generate a class-agnostic localization map. (b) In contrast, the proposed \nn~uses multiple class tokens to learn interactions between class tokens and patch tokens. The learned class-to-patch attentions of different class tokens can produce class-specific object localization maps. }
\vspace{-12pt}
\label{fig:teaser}
\end{figure}

%% file: sec/2_related.tex
\input{fig/overview} 

\section{Related works}
\label{sec:related}

\subsection{Weakly supervised semantic segmentation} 
Most existing WSSS approaches rely on Class Activation Mapping~\cite{zhou2016learning} to extract object localization maps from CNNs. The raw CAM maps are incomplete with coarse boundaries and thus unable to provide sufficient supervision to the learning of semantic segmentation networks. To tackle this problem, specific segmentation losses have been proposed to cater for deficient segmentation supervision, including SEC loss~\cite{kolesnikov2016seed}, CRF loss~\cite{tang2018regularized, zhang2019reliability} and contrastive loss~\cite{ke2021universal}. In addition, a number of studies have focused on improving the pseudo segmentation labels obtained from CAM maps.
These methods can be categorized as follows:

\par\noindent\textbf{Generating high-quality CAM maps.} 
A few methods developed heuristic strategies, such as ``Hide \& Seek"~\cite{singh2017hide} and Erasing~\cite{wei2017object}, applied either on images~\cite{zhang2021complementary, li2018tell} or feature maps~\cite{lee2019ficklenet, hou2018self} to drive the network to learn novel object patterns. Prior works also exploited sub-categories~\cite{chang2020weakly} and cross-image semantics~\cite{fan2020cian, sun2020mining, li2021group} to localize more accurate object regions.
To address the limitation of the standard classification objective loss function, regularization losses~\cite{wang2020self, zhang2020splitting} have been proposed to guide the network to discover more object regions. 
Moreover, several other works~\cite{wei2018revisiting, xu2021atrous} addressed the problem of the limited receptive field of standard image classification CNNs by introducing dilated convolutions, to encourage the propagation of discriminative activations to their surroundings. 

\par\noindent\textbf{Refining CAM maps with affinity learning.} Several works focused on learning pairwise semantic affinities to refine the CAM maps. Ahn~\etal~\cite{ahn2018learning} proposed AffinityNet to learn the affinities between adjacent pixels from the reliable seeds of the raw CAM maps. The learned AffinityNet can predict an affinity matrix to propagate the CAM maps via random walk. Similarly, Wang~\etal~\cite{wang2020weakly} also learned a pairwise affinity network using the confident pixels from the segmentation results. In~\cite{wang2020self,zhang2021complementary}, the affinity is directly learned from the feature maps of the classification network to refine the CAM maps. In addition, Xu~\etal~\cite{xu2021leveraging} proposed a cross-task affinity, which is learned from the saliency and segmentation representations in a weakly supervised multi-task framework.

In contrast to previous WSSS methods, which are all based on CNNs, we propose a transformer based model to extract class-specific object localization maps. 
We exploit the transformer attention map from the self-attention mechanism to generate object localization maps. 


\subsection{Transformers for visual tasks}
Transformers~\cite{vaswani2017attention}, were originally designed to model long-range dependencies of long sequences in the field of NLP. Recently, transformer models have been adapted to accommodate a wide variety of vision tasks~\cite{khan2021transformers}, such as image classification~\cite{dosovitskiy2020image}, saliency detection~\cite{liu2021visual} and semantic segmentation~\cite{ranftl2021vision}, achieving promising performances. The first transformer based vision model, ViT~\cite{dosovitskiy2020image}, splits an image into patches and transforms them into a sequence of tokens. These tokens are then forwarded into multiple stacked self-attention~\cite{vaswani2017attention} based layers, enabling each patch to have a global receptive field.

Caron~\etal~\cite{caron2021emerging} adapted self-supervised methods to ViT and observed that the attentions of the class token on patches contain information about the semantic layout of the scenes.
However, the one-to-one mapping between the attention and the class was not established in~\cite{caron2021emerging}. Besides,
their findings on the transformer attention have not been extended to weakly supervised learning.

Another related work, TS-CAM~\cite{gao2021ts}, adapts a CAM module to ViT. 
However, TS-CAM only leverages the class-agnostic attention maps of ViT, while the proposed method exploits class-specific localization maps from the transformer attention. Moreover, the proposed multi-class token transformer framework is shown to better complement the CAM mechanism than the original ViT, generating better object localization maps than TS-CAM (see \Table{abla_mct_pgt}).

%% file: fig/overview.tex
\begin{figure*}
\begin{center}
\includegraphics[width=.95\textwidth]{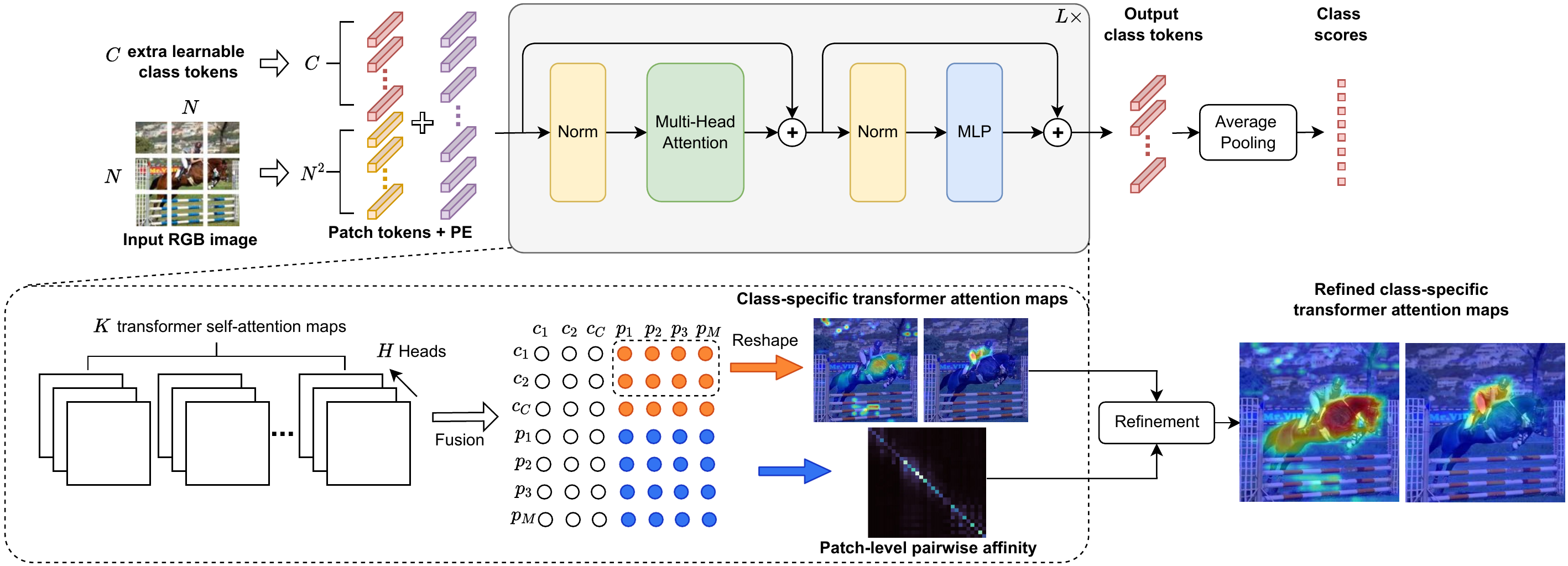}
\vspace{-15pt}
\end{center}
\caption{An overview of the proposed multi-class token transformer (\nn-V1). 
It first splits and transforms an input RGB image into a sequence of patch tokens. We propose to learn $C$ extra class tokens, where $C$ is the number of classes. The $C$ class tokens are concatenated with patch tokens, with added position embeddings (\textbf{PE}), which then go through consecutive $L$ transformer encoding layers. Finally, the output $C$ class tokens are used to produce class scores via average pooling. We aggregate the transformer attentions from the last $K$ layers and multiple heads to generate a final attention map, from which we can extract class-specific object localization maps and a patch-level pairwise affinity map from the class-to-patch and the patch-to-patch attentions, respectively. The patch-level pairwise affinity can be used to refine the class-specific transformer attention maps to produce improved object localization maps.}
\label{fig:overview}
\vspace{-12pt}
\end{figure*}

%% file: sec/3_method.tex
\section{Multi-class Token Transformer}

\subsection{Overview}
We propose a novel purely transformer-based framework (\nn-V1) to exploit class-specific object localization maps from the transformer attention. 
The overall architecture of \nn-V1 is shown in Figure~\ref{fig:overview}.
An input RGB image is first split into non-overlapping patches, which are then transformed into a sequence of patch tokens. In contrast to conventional transformers, which only use one class token, we propose to use multiple class tokens. 
These class tokens are concatenated with patch tokens, embedding position information, to form the input tokens of the transformer encoder. Several transformer blocks are used in the transformer encoder to extract features for both patch tokens and class tokens.
We apply average pooling on the output class tokens from the last layer to generate class scores, instead of using a Multi-Layer Perception (MLP) as in conventional transformers, for classification prediction.

At training time,
to ensure that different class tokens can learn different class-specific representations, we adopt the class-aware training strategy detailed in Section~\ref{sec:3.2}.
A classification loss is computed between the class scores directly produced by class tokens and the ground-truth class labels. This thus enables a strong connection between each class token and the corresponding class label. 
At test time, we can extract class-specific localization maps from the class-to-patch attention in the transformer. 
We further aggregate the attention maps from multiple layers to utilize complementary information learned from different transformer layers.  
Moreover, a patch-level pairwise affinity can be extracted from the patch-to-patch attentions, to further refine the class-to-patch attentions, leading to significantly improved class-specific localization maps. The class-specific localization maps are used as the seed to generate pseudo labels to supervise the segmentation model.

\subsection{Class-Specific Transformer Attention Learning}
\label{sec:3.2}

\par\noindent\textbf{Multi-class token structure design.} 
Consider an input image, it is split into $N\times N$ patches, which are then transformed into a sequence of patch tokens $\mathbf{T}_{p} \in \mathbb{R}^{M\times D}$, where $D$ is the embedding dimension, $M=N^2$. We propose to learn $C$ class tokens $\mathbf{T}_{cls} \in \mathbb{R}^{C\times D}$, where $C$ is the number of classes. The $C$ class tokens are concatenated with patch tokens, with added position embeddings to form the input tokens $\mathbf{T}_{in} \in \mathbb{R}^{(C+M)\times D}$ to the transformer encoder. The transformer encoder has $L$ consecutive encoding layers, each of which consists of a Multi-Head Attention (MHA) module, a MLP, and two LayerNorm layers applied before the MHA and the MLP, respectively.

\par\noindent\textbf{Class-specific multi-class token attention.} 
We use the standard self-attention layer to capture the long-range dependencies between tokens. More specifically, 
we first normalize the input token sequence and transform it to a triplet of  $\mathbf{Q} \in \mathbb{R}^{(C+M)\times D}$, $\mathbf{K} \in \mathbb{R}^{(C+M)\times D}$ and $\mathbf{V} \in \mathbb{R}^{(C+M)\times D}$, through linear layers~\cite{dosovitskiy2020image}. We employ the Scaled Dot-Product Attention~\cite{vaswani2017attention} mechanism to compute the attention values between the queries and keys. Each output token is a weighted sum of all tokens using the attention values as weights, formulated as:
\begin{equation}
    \mathrm{Attention}(\mathbf{Q}, \mathbf{K}, \mathbf{V}) = \mathrm{softmax}(\mathbf{Q}\mathbf{K}^\top/\sqrt{D})\mathbf{V},
\end{equation}
where we can obtain a token-to-token attention map $\mathbf{A}_{t2t} \in \mathbb{R}^{(C+M)\times (C+M)}$ and $\mathbf{A}_{t2t} = \mathrm{softmax}(\mathbf{Q}\mathbf{K}^\top/\sqrt{D})$.

From the global pairwise attention map $\mathbf{A}_{t2t}$, we can extract the class attentions to patches $\mathbf{A}_{c2p}\in \mathbb{R}^{C\times M}$, \ie, class-to-patch attention, where $\mathbf{A}_{c2p} = \mathbf{A}_{t2t}[1:C,C+1:C+M]$, as illustrated by the matrix with yellow dots in \Figure{overview}. Each row represents the attention scores of a specific class to all patches. Leveraging these attention vectors, with the original spatial positions of all patches, can produce $C$ class-relevant localization maps. 
We can extract class-relevant localization maps from each transformer encoding layer. Given that higher layers learn more high-level discriminative representations (while earlier layers capture more general and low-level visual information), we propose to fuse the class-to-patch attentions from the last $K$ transformer encoding layers, to explore a good trade-off between precision and recall on the generated object localization maps.  This process is formulated as:
\begin{equation}
\setlength{\abovedisplayskip}{3pt}
\setlength{\belowdisplayskip}{3pt}
    \hat{\mathbf{A}}_{mct} = \frac{1}{K}\sum_{l}^{K}\hat{\mathbf{A}}^{l}_{mct},
\end{equation}
where $\hat{\mathbf{A}}^{l}_{mct}$ is the class-specific transformer attention extracted from the $l^{th}$ transformer encoding layer of the proposed \nn-V1. The fused maps $\hat{\mathbf{A}}_{mct}$ are further normalized by the min-max normalization method along the two spatial dimensions, to generate the final class-specific object localization maps $\mathbf{A}_{mct} \in \mathbb{R}^{C\times N\times N}$. Detailed results on how to choose $K$ can be found in \Figure{lineplot}.

\input{fig/overview_c}

\par\noindent\textbf{Class-specific attention refinement.} The pairwise affinity is often used in prior works~\cite{ahn2018learning,wang2020weakly,xu2021leveraging} to refine object localization maps. It usually requires an additional network or extra layers to learn an affinity map.
In contrast, we propose to extract a pairwise affinity map from the patch-to-patch attention of the proposed \nn, without additional computations nor supervision.
This is achieved by extracting the patch-to-patch attentions $\mathbf{A}_{p2p}\in \mathbb{R}^{M\times M}$, where $\mathbf{A}_{p2p} = \mathbf{A}_{t2t}[C+1:C+M,C+1:C+M]$, as illustrated by the matrix with blue dots in \Figure{overview}. The patch-to-patch attentions are reshaped to a 4D tensor $\hat{\mathbf{A}}_{p2p}\in \mathbb{R}^{N\times N\times N\times N}$.
The extracted affinity is used to further refine the class-specific transformer attention. This process is formulated as:
\begin{equation}
\setlength{\abovedisplayskip}{3pt}
    \mathbf{A}_{mct\_ref}(c,i,j) = \sum_{k}^{N}\sum_{l}^{N}\hat{\mathbf{A}}_{p2p}(i,j,k,l)\cdot \mathbf{A}_{mct}(c,k,l),
    \label{attn_refine}
\end{equation}
where $\mathbf{A}_{mct\_ref}\in \mathbb{R}^{C\times N\times N}$ is the refined class-specific localization map. 
As shown in \Table{abla_mct_pgt} and \Figure{mct_cam}, using the patch-level pairwise affinity for refinement leads to better object localization maps with improved appearance continuity. This was not observed in the prior work~\cite{gao2021ts}.

\par\noindent\textbf{Class-aware training.} In contrast to conventional transformers which use the single class token from the last layer to perform classification prediction through a MLP, we have multiple class tokens $\mathbf{T}_{cls} \in \mathbb{R}^{C\times D}$,
and we need to ensure that different class tokens can learn different class-discriminative information. To this end,
we apply average pooling on the output class tokens to produce class scores:
\begin{equation}
\setlength{\abovedisplayskip}{2pt}
\setlength{\belowdisplayskip}{1pt}
\mathbf{y}(c) = \frac{1}{D}\sum_{j}^{D}\mathbf{T}_{cls}(c,j),
\label{eq:cls_token_pred}
\end{equation}
where  $\mathbf{y} \in \mathbb{R}^{C}$ is the class prediction and $c\in{1,2,...,C}$. $\mathbf{T}_{cls}(c,j)$ denotes an element in $\mathbf{T}_{cls}$, \ie, the $j^{th}$ feature of the $c^{th}$ class token. We finally compute a multi-label soft margin loss between the class score $\mathbf{y}(c)$ for the class $c$ and its ground-truth label. This provides each class token with strong and direct class-aware supervision, making each class token be able to capture class-specific information.

\subsection{Complementarity to Patch-Token CAM}
\label{sec:3.3}

\input{fig/segresults}
We integrate a CAM module~\cite{zhang2018adversarial,gao2021ts, zhou2016learning} into the proposed multi-class token transformer framework, as shown in Figure~\ref{fig:overview_c}, constructing an extended model, coined as \nn-V2. More specifically, given a sequence of output tokens from the transformer encoder $\mathbf{T}_{out} \in \mathbb{R}^{(C+M)\times D}$, we divide it into the output class tokens $\mathbf{T}_{out\_cls} \in \mathbb{R}^{C\times D}$ and the output patch tokens $\mathbf{T}_{out\_pat} \in \mathbb{R}^{M\times D}$. The patch tokens are then reshaped and forwarded to a convolutional layer with $C$ output channels, producing a 2D feature map $\mathbf{F}_{out\_pat} \in \mathbb{R}^{N\times N\times C}$. 
$\mathbf{F}_{out\_pat}$ is finally transformed to class predictions through a global average pooling (GAP) layer. In addition, we also use the output class tokens to produce class scores (see Eq.~\eq{cls_token_pred}). The total loss is the sum of two multi-label soft margin losses computed between the image-level ground-truth labels and the class predictions respectively from the class tokens and the patch tokens as follows:
\begin{equation}
\setlength{\abovedisplayskip}{3pt}
    \mathcal{L}_{total} = \mathcal{L}_{cls-class} + \mathcal{L}_{cls-patch}.
\end{equation}

\par\noindent\textbf{Combining PatchCAM and class-specific transformer attention.} At test time, patch token-based CAM (thereafter called PatchCAM) maps can be extracted from the last convolutional layer. We extract the PatchCAM maps $\mathbf{A}_{pCAM}$ with $\mathbf{A}_{pCAM}\in \mathbb{R}^{N\times N\times C}$, by applying the min-max normalization on the feature map $\mathbf{F}_{out\_pat}$. The extracted PatchCAM maps are then combined with the proposed class-specific transformer attention maps to produce fused object localization maps $\mathbf{A}$ through an element-wise multiplication operation:
\begin{equation}
\setlength{\abovedisplayskip}{3pt}
\setlength{\belowdisplayskip}{3pt}
    \mathbf{A} = \mathbf{A}_{pCAM} \circ \mathbf{A}_{mct},
\end{equation}
where $\circ$ denotes the Hadamard product. 

\par\noindent\textbf{Class-specific object localization map refinement.}
Similar to the attention refinement mechanism proposed in MCTformer-V1 (see Eq.~\ref{attn_refine}), we can also extract the patch-to-patch attention map from \nn-V2 as a patch-level pairwise affinity to refine the fused object localization maps as follows:
\begin{equation}
\setlength{\abovedisplayskip}{3pt}
\setlength{\belowdisplayskip}{3pt}
    \mathbf{A}_{ref}(c,i,j) = \sum_{k}^{N}\sum_{l}^{N}\hat{\mathbf{A}}_{p2p}(i,j,k,l)\cdot \mathbf{A}(c,k,l).
\end{equation}
\nn-V2~provides an effective transformer-based framework in which the CAM method can flexibly and robustly adapt to multi-label images.
By applying the classification loss on class predictions from both class tokens and patch tokens,
the strong consistency between these two types of tokens can be enforced to improve the model learning. The intuition is mainly two-fold. First, this consistency constraint can be regarded as an auxiliary supervision to guide the learning of more effective patch representations. Second, the strong pairwise interaction (\ie~message passing) between the patch tokens and the multiple class tokens can also lead to more representative patch tokens, thus producing more class-discriminative PatchCAM maps, compared to only using one class token as in TS-CAM~\cite{gao2021ts}.

%% file: fig/overview_c.tex
\begin{figure*}
\begin{center}
\includegraphics[width=.98\textwidth]{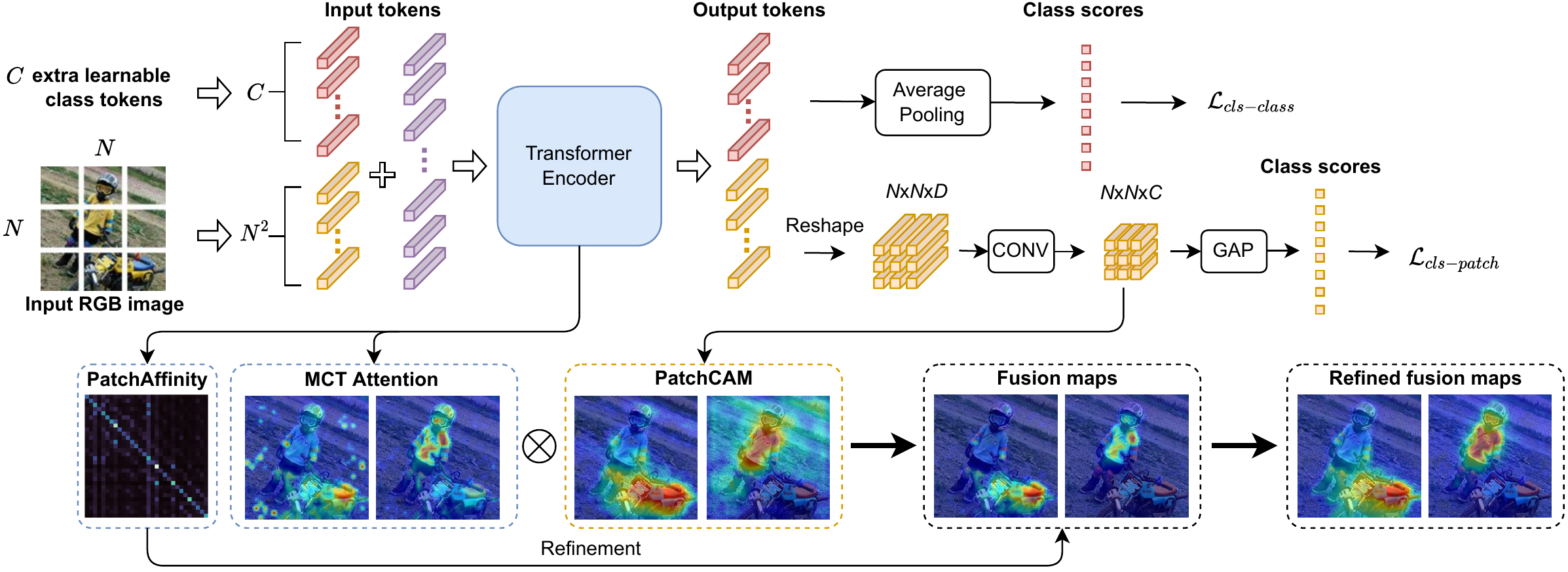}
\vspace{-15pt}
\end{center}
\caption{An overview of the proposed \nn-V2. We introduce a CAM module into the proposed \nn-V1. More specifically, the CAM module is composed of a convolutional layer and a global average pooling (GAP) layer. It takes the reshaped output patch tokens from the last transformer encoding layer as inputs, and outputs class scores. As for \nn-V1, we also use the output class tokens to produce class scores. The whole model is thus optimized by two classification losses applied on separately two types of class predictions. At the inference time, we fuse the class-specific transformer attentions (MCT Attention) and the PatchCAM maps. The results are further refined by the patch affinity extracted from the patch-to-patch transformer attentions to produce the final object localization maps.
}
\label{fig:overview_c}
\vspace{-12pt}
\end{figure*}

%% file: fig/segresults.tex
\begin{figure*}
\begin{center}
\includegraphics[width=.85\textwidth]{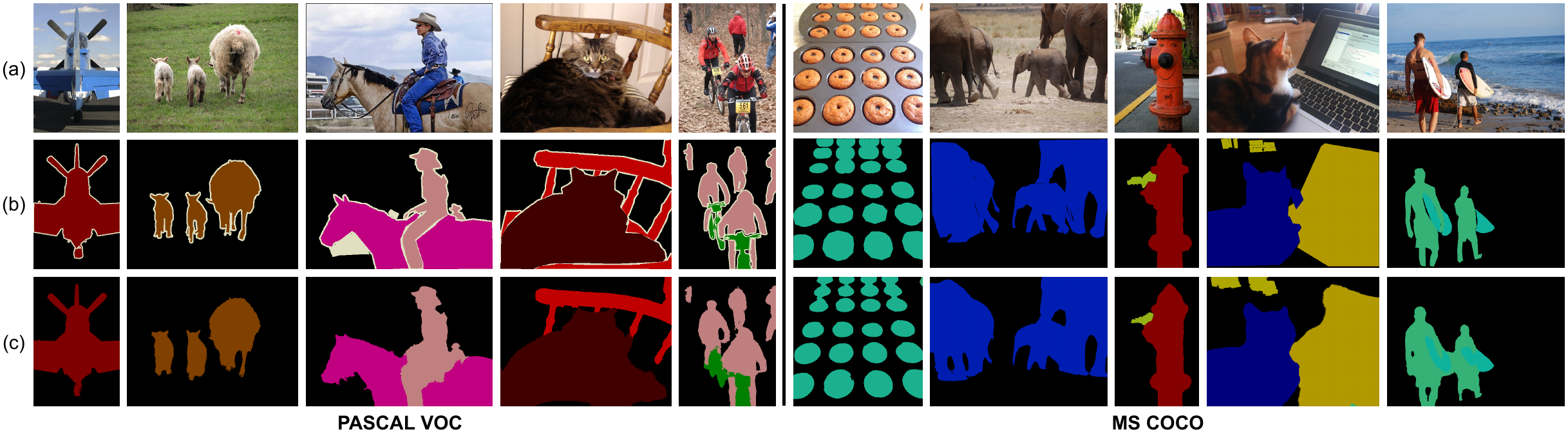}
\vspace{-18pt}
\end{center}
\caption{Qualitative segmentation results on the PASCAL VOC and MS COCO \textit{val} sets. (a) Input. (b) Ground-truth. (c) Ours.}
\label{fig:segresults}
\vspace{-12pt}
\end{figure*}

%% file: sec/4_results.tex
\section{Experiments}
\subsection{Experimental Settings}
\par\noindent\textbf{Datasets.} We evaluated the proposed approach on two datasets, \ie, PASCAL VOC 2012~\cite{everingham2010pascal} and MS COCO 2014~\cite{lin2014microsoft}. \textbf{PASCAL VOC} has three subsets, \ie, training (train),  validation (val) and test sets, each containing 1,464, 1,449, and 1,456 images, respectively. It has 20 object classes and one background class for the semantic segmentation task. Following prior works~\cite{chang2020weakly,wang2020self,lee2021anti,su2021context,zhang2021complementary,xu2021leveraging}, an augmented set of 10,582 images, with additional data from~\cite{hariharan2011semantic}, was used for training. \textbf{MS COCO} uses 80 object classes and one background class for semantic segmentation. Its training and validation sets contain 80K and 40K images, respectively. Note that we only used \emph{image-level} ground-truth labels from these datasets during training. 
\input{tab/abla_pgt_voc}

\par\noindent\textbf{Evaluation metrics.}
In line with previous works~\cite{lee2021anti}, we use the mean Intersection-over-Union (mIoU) to evaluate 
the semantic segmentation performance on the \textit{val} set, of the two benchmarks. We obtained the semantic segmentation results on the PASCAL VOC \textit{test} set from the official PASCAL VOC online evaluation server.
\par\noindent\textbf{Implementation details.}
We built the proposed \nn~using the DeiT-S Backbone~\cite{touvron2021training, gao2021ts} pre-trained on ImageNet~\cite{deng2009imagenet}. More specifically, we used the pre-trained class token in DeiT-S to initialize the proposed multiple class tokens. We followed the data augmentation and default training parameters provided in~\cite{touvron2021training, gao2021ts}. Training images are resized to $256\times256$ and then cropped into $224\times224$. For semantic segmentation, we followed prior works~\cite{ahn2018learning, zhang2019reliability, xu2021leveraging, zhang2021complementary} to use the ResNet38~\cite{wu2019wider} based Deeplab V1. 
At test time, we used multi-scale testing and CRFs with the hyper-parameters suggested in~\cite{chen2014semantic} for post-processing.

\input{tab/sota_voc}

\subsection{Comparison with State-of-the-arts}
\par\noindent\textbf{PASCAL VOC.} We followed the common practice~\cite{chang2020weakly,wang2020self,lee2021anti,su2021context,zhang2021complementary} to apply PSA~\cite{ahn2018learning} on the proposed object localization maps (seed) to generate pseudo semantic segmentation ground-truth labels (mask) on the \textit{train} set. As shown in~\Table{seed}, the proposed method performs better than existing works by large margins on both the initial seed and the pseudo ground-truth mask, better than the best initial seed~\cite{zhang2021complementary} by 4.3\%. \Table{sota_res38} shows that the proposed MCTformer achieves segmentation results (mIoUs) of 71.9\% and 71.6\% on the \textit{val} and \textit{test} sets, respectively. The proposed \nn~performs significantly better than all the existing methods using only image-level labels. In particular, \nn~can even achieve comparable or better results compared to the methods using additional saliency maps. \Figure{segresults} (a) shows that the segmentation model trained with our pseudo labels can produce accurate and complete object contours in various challenging scenes.
\input{tab/sota_coco}
\par\noindent\textbf{MS COCO.} 
\Table{coco} shows that the proposed method achieves a segmentation mIoU of 42.0\%, surpassing the recent methods by a large margin. Notably, from~\Table{coco}, we observe that several methods using additional saliency maps obtain inferior performance compared to recent methods only using image-level labels. This reveals the limitation of relying on pre-trained saliency models, which may perform poorly on challenging datasets. Several qualitative segmentation results are shown in~\Figure{segresults} (b).
\input{tab/complexity}
\par\noindent\textbf{Model complexity.} We compared the model complexity of the proposed MCTformer with a commonly used CNN model for generating object localization maps~\cite{ahn2018learning,wang2020self,zhang2021complementary}, \ie, ResNet38~\cite{wu2019wider}, in terms of the number of parameters and multiply-add calculations (MACs). \Table{complexity} shows that the complexity of the proposed DeiT-S~\cite{touvron2021training} based method is significantly smaller than that of ResNet38 based methods.

\subsection{Ablation Studies}
\label{sec:abla}
\input{fig/mct_cam}
\input{tab/abla_variants}
\input{tab/abla_variants_seg}
\input{tab/abla_arch}
\input{fig/abla_layers}
\par\noindent\textbf{Effect of multiple class-specific tokens}.
In the conventional ViT, the class token attention only indicates a class-agnostic localization map. TS-CAM~\cite{gao2021ts} applies CAM on the output patch tokens of ViT to obtain class-specific localization maps. Following their official implementation, the generated object localization maps by TS-CAM on the PASCAL VOC \textit{train} set obtained an mIoU of 29.9\%, as shown in \Table{abla_mct_pgt}. We simply added a ReLU layer on their PatchCAM maps (\ie, TS-CAM$^\ast$), yielding a large improvement of 11.4\%. 
In comparison, the proposed baseline method, \textit{i.e.,} the class-specific transformer attention maps of the multiple class-specific tokens in the proposed \nn-V1, attains an mIoU of 47.2\%, outperforming TS-CAM$^\ast$ by a significant margin of 5.9\%. 
This demonstrates the effectiveness of the proposed transformer attention based class-specific localization maps.
\par\noindent\textbf{Complementarity of PatchCAM and the proposed class-specific transformer attention}. \Table{abla_mct_pgt} shows that the object localization maps generated by \nn-V2 with a standard CAM module, obtain an mIoU of 58.2\%. This can be further improved to 61.7\% by using the patch-level pairwise affinity for refinement.
As shown in \Figure{mct_cam}e, the class-specific transformer attention can effectively localize objects while with low responses and noises. In contrast, the PatchCAM maps (\Figure{mct_cam}f) show high responses on object regions, while they also have more background pixels around the objects activated. The fusion of these two leads to clearly improved localization maps which only activate object regions, with significantly reduced background noises (\Figure{mct_cam}g). These class-specific localization maps confirm remarkably superior performance of our proposed model compared to TS-CAM~\cite{gao2021ts} (\Figure{mct_cam}b) that shows sparse and low object responses in most cases.

\par\noindent\textbf{Effect of patch affinity.} As shown in \Table{abla_mct_pgt} and \Table{abla_mct_seg}, by applying the learned patch-to-patch attention as a patch-level pairwise affinity to refine the object localization maps from \nn-V1, the pseudo segmentation label maps can be improved by 8\%, and accordingly, the segmentation performance is also improved by a gain of 3.2\%. \nn-V2 yields consistent improvements in terms of the quality of generated pseudo labels and the segmentation performance, compared to the variants that do not use patch affinity. The visualization results in \Figure{mct_cam} (d) and (h) show that the refined object localization maps appear more complete with smoother object contours. This further demonstrates the great benefits of our method in generating an effective patch affinity without additional computation.

\par\noindent\textbf{Different class prediction methods.} We evaluated the effect of the different strategies used to produce class scores on the generated class-specific transformer attention maps.
As shown in~\Table{average-pool}, max pooling has the worst performance for class-specific localization with an mIoU of only 26.8\%, while using a fully connected layer for a linear projection yields an improved mIoU of 41.5\%. The average pooling produces the best performance with an mIoU of 47.2\%. These results confirm our initial design motivation.
Specifically, involving extra parameters within a fully-connected layer may increase the complexity of learning the model for discriminative localization. 
In contrast to the max pooling which only needs to attend to the most relevant patch, the average pooling can encourage class tokens to attend to more relevant patches, which is beneficial to learn better spatial context for localization.

\par\noindent\textbf{Number of layers for attention fusion.} We evaluated the quality of object localization maps by fusing attention maps of different class tokens from multiple transformer encoding layers in the proposed~\nn-V1.~Following~\cite{wang2020self}, we use three evaluation metrics, \ie, false positives (FP), false negatives (FN), and mIoU, in which larger FP and FN values indicate increased over-activated and under-activated regions, respectively. 
As shown in~\Figure{lineplot}, aggregating information from more layers produces object localization maps which tend to be over-activated. This indicates that 
the early layers produce more generic low-level representations which may not be very helpful for high-level semantic localization. By decreasing the number of layers, the generated object localization maps become more discriminative at the cost of a lower activation coverage. The results reported in~\Figure{lineplot} suggest that fusing the attentions from the last three layers can yield the best pseudo segmentation ground-truth labels
(mIoU of 47.2\%). 

%% file: tab/abla_pgt_voc.tex

\begin{table}[t]
\caption{Evaluation of the initial seed (Seed) and the corresponding pseudo segmentation ground-truth mask (Mask) in terms of mIoU (\%) on the PASCAL VOC \textit{train} set. }
\vspace{-8pt}
\label{tab:seed}
\small
\centering
\resizebox{0.7\linewidth}{!}
{\begin{tabular}{lcc}
\toprule 
Method            & Seed & Mask                   \\ \midrule
PSA (CVPR18) \cite{kolesnikov2016seed} & 48.0 & 61.0 \\
Chang \etal (CVPR20) \cite{chang2020weakly} & 50.9&63.4 \\
SEAM (CVPR20) \cite{wang2020self} &55.4& 63.6\\
AdvCAM (CVPR21) \cite{lee2021anti} &55.6& 68.0 \\
CDA (ICCV21) \cite{su2021context} &55.4&63.4 \\
Zhang \etal (ICCV21) \cite{zhang2021complementary} &57.4& 67.8 \\
 \midrule
\textbf{MCTformer} (Ours) & \textbf{61.7} & \textbf{69.1} \\
  \bottomrule 
\end{tabular}
}
\vspace{-18pt}
\end{table}

%% file: tab/sota_voc.tex
\begin{table}[t]
\caption{Performance comparison of WSSS methods in terms of mIoU (\%) on the PASCAL VOC 2012 \textit{val} and \textit{test} sets using different segmentation backbones. Sup.: supervision. I: image-level ground-truth labels. S: off-the-shelf saliency maps. 
\label{tab:sota_res38}}
\vspace{-8pt}
\centering
\small
\resizebox{1.0\linewidth}{!}{
\begin{tabular}{lcccc}
\toprule
Method            & Backbone  & Sup.          & Val            & Test           \\ \midrule
  CIAN (AAAI20)  \cite{fan2020cian}    &         ResNet101     &     I+S&  64.3         &  65.3 \\
ICD (CVPR20) \cite{fan2020learning}    &          ResNet101          &   I+S&    67.8         &    68.0           \\ 
 Zhang \etal (ECCV20) \cite{zhang2020splitting}   &       ResNet50  &   I+S&   66.6         &   66.7    \\  
 Sun \etal (ECCV20) \cite{sun2020mining}   &          ResNet101        &I+S&        66.2          &      66.9        \\
 EDAM (CVPR21) \cite{wu2021embedded} & ResNet101 & I+S & 70.9 & 70.6 \\
 EPS (CVPR21) \cite{lee2021railroad} & ResNet101 & \textbf{I+S} & \textbf{71.0}&\textbf{71.8} \\
 Yao \etal (CVPR21) \cite{yao2021non} & ResNet101 & I+S &68.3&68.5 \\
 AuxSegNet (ICCV21) \cite{xu2021leveraging} & ResNet38 & I+S &69.0&68.6 \\ \hline
  Zhang \etal (AAAI20) \cite{zhang2019reliability} & ResNet38 & I&62.6 & 62.9 \\
    Luo \etal (AAAI20)  \cite{luo2020learning}  &     ResNet101         &I     &       64.5       & 64.6 \\
Chang \etal (CVPR20) \cite{chang2020weakly}  &            ResNet101    &I     &      66.1        &    65.9\\ 
Araslanov \etal (CVPR20) \cite{araslanov2020single}    &          ResNet38    &I     &      62.7        &   64.3       \\ 
SEAM (CVPR20) \cite{wang2020self} & ResNet38 &I& 64.5 & 65.7 \\
 BES (ECCV20) \cite{chen2020weakly} & ResNet101 & I &65.7&66.6 \\
 CONTA (NeurIPS20) \cite{zhang2020causal}   &          ResNet38         &  I&    66.1       &    66.7            \\  
AdvCAM (CVPR21) \cite{lee2021anti} & ResNet101 & I &68.1 &68.0 \\
 ECS-Net (ICCV21) \cite{sun2021ecs} & ResNet38 & I &66.6 & 67.6 \\
 Kweon \etal (ICCV21) \cite{kweon2021unlocking} & ResNet38 & I &68.4&68.2 \\
 CDA (ICCV21) \cite{su2021context} & ResNet38 & I &66.1&66.8 \\
 Zhang \etal (ICCV21) \cite{zhang2021complementary} & ResNet38 & I & 67.8&68.5 \\
 \midrule
  \textbf{\nn} (Ours) & ResNet38 &\textbf{I}&    \textbf{71.9} &  \textbf{71.6}\\
  \bottomrule
\end{tabular}
}
\vspace{-15pt}
\end{table}

%% file: tab/sota_coco.tex
\begin{table}[t]
\caption{Performance comparison of WSSS methods in terms of mIoU(\%) on the MS COCO \textit{val} set. }
\vspace{-8pt}
\label{tab:coco}
\small
\centering
\begin{tabular}{lccc}
\toprule 
Method            & Backbone  &Sup.          & Val                       \\ \midrule
EPS (CVPR21) \cite{lee2021railroad} & ResNet101 & \textbf{I+S} & \textbf{35.7} \\
AuxSegNet (ICCV21) \cite{xu2021leveraging} & ResNet38 & I+S & 33.9 \\ \midrule
Wang \etal (IJCV20) \cite{wang2020weakly}&VGG16&I&27.7\\
    Luo \etal (AAAI20)  \cite{luo2020learning}  &     VGG16              &  I&    29.9 \\
SEAM (CVPR20) \cite{wang2020self} & ResNet38 & I&31.9 \\
CONTA (NeurIPS20) \cite{zhang2020causal} & ResNet38 & I&32.8 \\
Kweon \etal (ICCV21) \cite{kweon2021unlocking} & ResNet38 & I & 36.4 \\
CDA (ICCV21) \cite{su2021context} & ResNet38 & I &33.2 \\
 \midrule
 \textbf{\nn} (Ours) & ResNet38 &\textbf{I }& \textbf{42.0} \\
  \bottomrule 
\end{tabular}
\vspace{-10pt}
\label{tab:segsota}
\end{table}

%% file: tab/complexity.tex
\begin{table}[t]
\caption{Complexity of models generating object localization maps. The proposed \nn~is based on DeiT-S~\cite{touvron2021training}.}
\vspace{-8pt}
\label{tab:complexity}
\small
\centering
\begin{tabular}{lcccc}
\toprule 
Model&Image size&\#Params (M)&MACs (G) \\ \midrule
ResNet38&$224\times224$&104.3&99.8 \\
\nn-V1&$224\times224$&21.7&4.6\\
\nn-V2&$224\times224$&21.8&4.7 \\
  \bottomrule 
\end{tabular}
\vspace{-10pt}
\end{table}

%% file: fig/mct_cam.tex
\begin{figure*}
\begin{center}
\includegraphics[width=.89\textwidth]{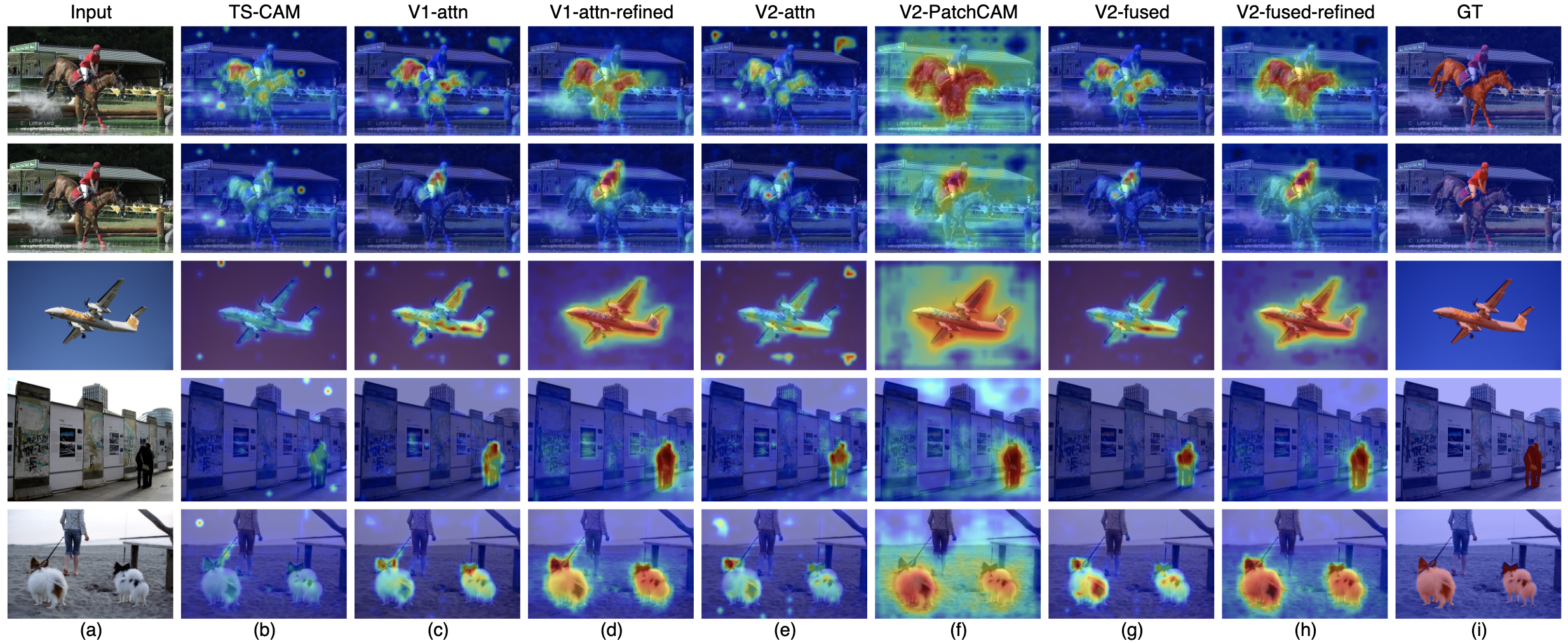}
\vspace{-15pt}
\end{center}
\caption{Visualization examples of different object localization maps from different methods: (b) TS-CAM~\cite{gao2021ts}; (c) V1-attn (the class-specific transformer attention from our \nn-V1); (d) V1-attn-refined (the refined class-specific transformer attention by the patch affinity from \nn-V1); (e) V2-attn (the class-specific transformer attention from \nn-V2); (f) V2-PatchCAM (the PatchCAM maps from \nn-V2). (g) V2-fused (the fused map of the class-specific transformer attention and the PatchCAM map from \nn-V2); (h) V2-fused-refined (the refined fusion map by the patch affinity from \nn-V2). (i) Ground-truth.}
\label{fig:mct_cam}
\vspace{-12pt}
\end{figure*}

%% file: tab/abla_variants.tex
\begin{table}[t]
\caption{Evaluation of different object localization maps in terms of mIoU(\%) on the PASCAL VOC \textit{train} set. }
\vspace{-8pt}
\label{tab:abla_mct_pgt}
\small
\centering
\resizebox{1.0\linewidth}{!}{
\begin{tabular}{lc}
\toprule 
Method           & mIoU                   \\ \midrule
TS-CAM~\cite{gao2021ts} &29.9\\
TS-CAM$^\ast$~\cite{gao2021ts} & 41.3 \\
\nn-V1 (Attention) & 47.2 \\
\nn-V1 (Attention + PatchAffinity) & 55.2 \\
\nn-V2 (Attention + PatchCAM) &58.2 \\
\nn-V2 (Attention + PatchCAM + PatchAffinity) & \textbf{61.7} \\ 
\bottomrule
\end{tabular}
}
\vspace{-15pt}
\end{table}

%% file: tab/abla_variants_seg.tex
\begin{table}[t]
\caption{Segmentation results using different object localization maps in terms of mIoU(\%) on the PASCAL VOC \textit{val} set. }
\vspace{-8pt}
\label{tab:abla_mct_seg}
\small
\centering
\resizebox{1.0\linewidth}{!}{
\begin{tabular}{lc}
\toprule 
Method           & mIoU                   \\ \midrule
TS-CAM$^\ast$~\cite{gao2021ts} & 49.7 \\
\nn-V1 (Attention) & 55.6 \\
\nn-V1 (Attention + PatchAffinity) & 58.8 \\
\nn-V2 (Attention + PatchCAM) &61.1 \\
\nn-V2 (Attention + PatchCAM + PatchAffinity) & \textbf{62.6} \\ 
\bottomrule
\end{tabular}
}
\vspace{-10pt}
\end{table}

%% file: tab/abla_arch.tex
\begin{table}[t]
\caption{Comparison of different methods for class prediction in \nn-V1 on the generated class-specific transformer attention in terms of mIoU(\%) on the PASCAL VOC \textit{train} set. }
\vspace{-8pt}
\label{tab:average-pool}
\small
\centering
\begin{tabular}{lccc}
\toprule 
          & Fully-connected  & Max-pooling & Average-pooling                   \\ \midrule
mIoU &41.5&26.8&\textbf{47.2}\\
  \bottomrule 
\end{tabular}
\vspace{-12pt}
\end{table}

%% file: fig/abla_layers.tex
\begin{figure}
\begin{center}
\includegraphics[width=.4\textwidth]{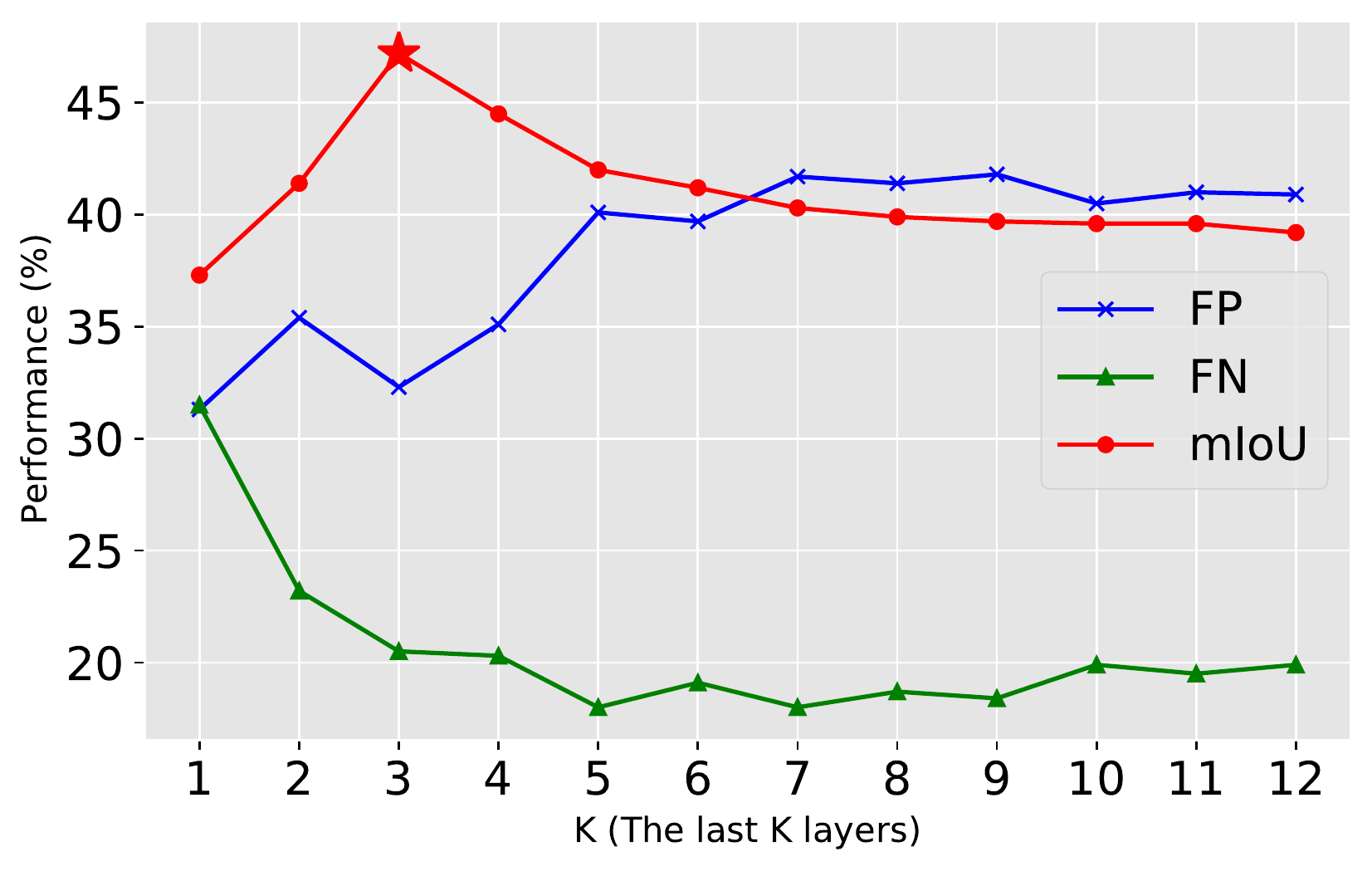}
\end{center}
\vspace{-15pt}
\caption{Evaluation of object localization maps generated by fusing the class token attentions from the last $K$ transformer layers in terms of false positives (FP), false negatives (FN) and mIoU.}
\label{fig:lineplot}
\vspace{-18pt}
\end{figure}

%% file: sec/5_conclusions.tex
\vspace{-5pt}
\section{Conclusions}
\vspace{-5pt}
This paper presents \nn, a simple yet effective transformer-based framework to produce class-specific object localization maps, and achieves state-of-the-art results on WSSS. 
We show that class-to-patch attention of different class tokens can discover class-specific localization information, while patch-to-patch attention can also learn effective pairwise affinities to refine the localization maps. 
Furthermore, we demonstrate that the proposed framework can seamlessly complement the CAM mechanism, 
leading to high-quality pseudo ground-truth labels for weakly supervised semantic segmentation. 

%% file: sec/X_supplementary.tex
\appendix

\setcounter{page}{1}

\twocolumn[
\centering
\Large
\textbf{Multi-class Token Transformer for Weakly Supervised Semantic Segmentation} \\
\vspace{0.5em}-Supplementary Material- \\
\vspace{1.0em}
] 
\appendix
\section{Implementation details}

\subsection{Training and testing of~\nn} 
To integrate the CAM module into the proposed \nn, we used a convolutional layer with $C$ kernels of $3\times3$, a stride of 1, and a padding of 1, where $C$ is the number of classes. We used the AdamW optimizer to train \nn~with a batch size of 64 and an initial learning rate of $5\times10^{-4}$. The number of training epochs was set to 60. At test time, we generated transformer attention maps by fusing attentions from multiple layers. More specifically, to generate the class-specific object localization maps, we aggregated the class-to-patch transformer attention maps from the last three layers, as detailed~in Figure 6 in Section 4.3.
To generate the patch-level pairwise affinity, we aggregated the patch-to-patch transformer attention maps from all the twelve transformer layers, as the pairwise affinity is class-agnostic and different transformer blocks can learn different-level similarities between tokens. By aggregating all patch-to-patch attentions, it can thus produce a more informative affinity map. For the aggregation method, we followed~\cite{gao2021ts} to first average attention maps from all the heads in each transformer layer, and then added up the averaged attention maps from all the selected layers, and finally performed a normalization to output a target attention map. For the evaluation of the generated class-specific object localization maps, we followed~\cite{wang2020self, lee2021anti} to report optimal results obtained when applying a range of thresholds to determine background pixels.


\subsection{Training and testing for semantic segmentation}
Following prior works~\cite{ahn2018learning,zhang2019reliability,wang2020self,zhang2021complementary}, we used ResNet38 based DeepLab-V1 as the segmentation model. For data augmentation, we used random scaling with a factor of $\pm 0.3$, random horizontal flipping, random cropping to size $321\times 321$. The polynomial learning rate decay was chosen with an initial learning rate of $7\times10^{-4}$ and a power of 0.9. We used the stochastic gradient descent (SGD) optimizer to train the segmentation network for 30 epochs with a batch size of 4. At test time, we used multi-scale testing, \ie, using inputs of multiple scales (0.5, 0.75, 1.0, 1.25, 1.5), and max-pooling for aggregating outputs, and the CRF with the default hyper-parameters suggested in~\cite{chen2014semantic} for post-processing.

\section{Additional quantitative results}
We reported the per-class IoU results on both the \textit{val} and \textit{test} sets of PASCAL VOC, and the \textit{val} set of MS COCO in~\Table{iou_voc} and~\Table{iou_coco}, respectively. These results show that the proposed \nn~outperforms other state-of-the-art methods on most object categories, which demonstrates the superior performance of the proposed method.

\section{Additional qualitative results}
More qualitative segmentation results on the PASCAL VOC and MS COCO \textit{val} sets are presented in~\Figure{supp_voc_seg} and~\Figure{supp_coco_seg}. We can observe that the segmentation model trained with the pseudo labels generated by the proposed method produces satisfactory segmentation results. The model can segment large-scale objects with clear boundaries, and segment small-scale objects with fine-grained details in various indoor and outdoor scenes.

\Figure{supp_patch_aff} shows examples of the learned affinity maps of selected points (marked by the green crosses) in the input images, the generated transformer attention maps and their refined results by using the learned patch-to-patch transformer attention of the proposed \nn-V1. The learned affinity map (see~\Figure{supp_patch_aff}b) represents the similarity of the selected patch to all patches in the image. The cold (blue) to warm (red) colors denote low to high attention scores. As shown in the first row of~\Figure{supp_patch_aff}b, the affinity map highlights almost the entire object region for the class ``dog". Although the object patch is not activated in the original transformer attention map (marked by the red square in the first row of~\Figure{supp_patch_aff}c), the learned affinity propagates the activations from similar regions to this patch, thus increasing its activation scores (marked by the red square in~\Figure{supp_patch_aff}d). 
More examples of the class-specific transformer attention maps and corresponding refined results by using the affinity maps from the proposed \nn-V1 are presented in~\Figure{supp_v1_cam}. These results show that the proposed \nn-V1 can effectively generate class-specific object localization maps from the transformer attentions. 
In addition, the patch-to-patch transformer attention of \nn-V1 is used as affinity maps, which can not only activate non-discriminative regions, but also refine object localization maps with noise filtering to produce more accurate object maps and boundaries. 

More examples of the generated class-specific object localization maps from \nn-V2 on PASCAL VOC and MS COCO are presented in~\Figure{supp_v2_cam} and~\Figure{supp_v2_cam_coco}, respectively. \Figure{supp_v2_cam}b shows the PatchCAM maps that are extracted from the transformed patch tokens. The global receptive field of the transformer is beneficial for CAM to localize a full context of large-scale objects (\eg, the ``plane" and the ``train" in the third and fourth rows), while it leads to over-activated localization maps for small-scale or irregular objects, such as the ``bird" and the ``plant" in the first and the last rows of~\Figure{supp_v2_cam}. In contrast, as shown in~\Figure{supp_v2_cam}c, the transformer attention usually allocates small values evenly to large-scale objects, due to the self-attention mechanism that all attention values of a class token are summed up to one. For small-scale or slim objects such as the ``bird" in the first row of \Figure{supp_v2_cam}, the proposed transformer attention can generate object localization maps with clear boundaries.
The fusion of these two complementary maps, \ie, the PatchCAM maps and the class-specific transformer attention maps, leads to significantly improved class-specific object localization maps with highly activated object regions and 
largely suppressed noise (\Figure{supp_v2_cam}d and \Figure{supp_v2_cam_coco}d). Applying the patch-level pairwise affinity on the fused maps from these two can generate further refined object localization maps (\Figure{supp_v2_cam}e and \Figure{supp_v2_cam_coco}e). 

\input{supp/tab/iou_seg_voc}
\input{supp/tab/iou_seg_coco}
\input{supp/fig/voc_seg}
\input{supp/fig/coco_seg}
\input{supp/fig/supp-patch-aff}
\input{supp/fig/supp-v1-cam}
\input{supp/fig/supp-v2-cam-voc}
\input{supp/fig/supp-v2-cam-coco}

%% file: supp/tab/iou_seg_voc.tex
\begin{table*}[t]
\caption{Per-class performance comparison with the state-of-the-art WSSS methods in terms of IoUs (\%) on PASCAL VOC. $^*$ denotes without post-processing.}
\vspace{-8pt}
\label{tab:iou_voc}
\small
\centering
\resizebox{\linewidth}{!}
{\begin{tabular}{lcccccccccccccccccccccc}
\toprule 
&bkg&plane&bike&bird&boat&bottle&bus&car&cat&chair&cow&table&dog&horse&mbk&person&plant&sheep&sofa&train&tv&mIoU \\
\midrule
 \multicolumn{23}{l}{Results on the \textit{val} set:} \\
SEAM (CVPR20)~\cite{wang2020self}&88.8& 68.5& 33.3& 85.7 &40.4& 67.3& 78.9& 76.3& 81.9 &29.1& 75.5& 48.1 &79.9& 73.8 &71.4& 75.2& 48.9 &79.8& 40.9& 58.2& 53.0 &64.5\\
BEC (ECCV20)~\cite{chen2020weakly}&88.9& 74.1 &29.8& 81.3& 53.3& 69.9 &\textbf{89.4} &\textbf{79.8}& 84.2& 27.9& 76.9 &46.6& 78.8 &75.9& 72.2 &70.4& 50.8& 79.4& 39.9 &65.3 &44.8& 65.7\\
AdvCAM (CVPR21)~\cite{lee2021anti}&90.0& 79.8& 34.1 &82.6 &\textbf{63.3}& 70.5& \textbf{89.4}& 76.0 &87.3& 31.4& 81.3 &33.1 &82.5& 80.8& 74.0& 72.9& 50.3& 82.3 &42.2& \textbf{74.1} &52.9& 68.1 \\
ECS-Net (ICCV21)~\cite{sun2021ecs}&89.8 &68.4 &33.4 &85.6& 48.6& 72.2 &87.4 &78.1& 86.8& \textbf{33.0}& 77.5& 41.6 &81.7& 76.9& 75.4 &75.6 &46.2& 80.7& 43.9& 59.8 &56.3& 66.6\\
CDA (ICCV21)~\cite{su2021context}&89.1& 69.7& 34.5& 86.4& 41.3& 69.2& 81.3 &79.5& 82.1 &31.1& 78.3& 50.8 &80.6 &76.1& 72.2& 77.6 &48.8 &81.2& 42.5& 60.6& 54.3& 66.1\\
Zhang~\etal~ (ICCV21)~\cite{zhang2021complementary}&89.9&75.1&32.9&87.8&60.9&69.5&87.7&79.5&89.0&28.0&80.9&34.8&83.4&79.7&74.7&66.9&\textbf{56.5}&82.7&44.9&73.1&45.7&67.8 \\
Kweon~\etal~ (ICCV21)~\cite{kweon2021unlocking}&90.2& \textbf{82.9}& 35.1& 86.8 &59.4 &70.6& 82.5& 78.1& 87.4 &30.1& 79.4& 45.9& 83.1& 83.4& 75.7& 73.4 &48.1& \textbf{89.3} &42.7& 60.4 &52.3 &68.4\\
\textbf{\nn}$^*$~(Ours)& 90.6&71.8&37.5&85.1&52.9&68.8&78.8&78.7&87.1&28.4&78.9&53.0&83.9&78.2&76.8&76.4&54.1&80.1&46.0&71.6&54.3&68.2\\
\textbf{\nn}~(Ours)&\textbf{91.9}&78.3&\textbf{39.5}&\textbf{89.9}&55.9&\textbf{76.7}&81.8&79.0&\textbf{90.7}&32.6&\textbf{87.1}&\textbf{57.2}&\textbf{87.0}&\textbf{84.6}&\textbf{77.4}&\textbf{79.2}&55.1&89.2&\textbf{47.2}&70.4&\textbf{58.8}&\textbf{71.9} \\
 \midrule
  \multicolumn{23}{l}{Results on the \textit{test} set:} \\
AdvCAM (CVPR21)~\cite{lee2021anti}& 90.1 &81.2 &33.6& 80.4 &52.4& 66.6& 87.1& 80.5& 87.2& 28.9& 80.1& 38.5& 84.0 &83.0& \textbf{79.5}& 71.9& 47.5&80.8&59.1& 65.4 &49.7 &68.0\\
Zhang~\etal~ (ICCV21)~\cite{zhang2021complementary}&90.4&79.8&32.9&\textbf{85.8}&52.9&66.4&\textbf{87.2}&\textbf{81.4}&87.6&28.2&79.7&50.2&82.9&80.4&78.9&70.6&51.2&83.4&55.4&\textbf{68.5}&44.6&68.5 \\
\textbf{\nn}$^*$~(Ours)&90.9&76.0&37.2&79.1&54.1&69.0&78.1&78.0&86.1&30.3&79.5&58.3&81.7&81.1&77.0&76.4&49.2&80.0&55.1&65.4&54.5&68.4 \\
\textbf{\nn}~(Ours)&\textbf{92.3}&\textbf{84.4}&\textbf{37.2}&82.8&\textbf{60.0}&\textbf{72.8}&78.0&79.0&\textbf{89.4}&\textbf{31.7}&\textbf{84.5}&\textbf{59.1}&\textbf{85.3}&\textbf{83.8}&79.2&\textbf{81.0}&\textbf{53.9}&\textbf{85.3}&\textbf{60.5}&65.7&\textbf{57.7}&\textbf{71.6}\\
  \bottomrule 
\end{tabular}
}
\end{table*}

%% file: supp/tab/iou_seg_coco.tex
\begin{table*}[t]
\caption{Per-class performance comparison with the state-of-the-art WSSS methods in terms of IoU(\%) on the MS COCO \textit{val} set.
\label{tab:iou_coco}}
\centering
\resizebox{0.9\linewidth}{!}{
\begin{tabular}{lccc|lccc}
\toprule 
\multirow{2}{*}{Class} &Luo~\etal&AuxSegNet&MCTformer&\multirow{2}{*}{Class}&Luo~\etal&AuxSegNet&MCTformer          \\ 
&(AAAI20)~\cite{luo2020learning}&(ICCV21)~\cite{xu2021leveraging}&(Ours)&&(AAAI20)~\cite{luo2020learning}&(ICCV21)~\cite{xu2021leveraging}&(Ours)\\
\midrule 
background &73.9&82.0&\textbf{82.4}&wine class&27.2&\textbf{32.1}&27.0 \\
person &48.7&\textbf{65.4}&62.6&cup &21.7&\textbf{29.3}&29.0\\
bicycle &45.0&43.0&\textbf{47.4}&fork &0.0&5.4&\textbf{13.9} \\
car &31.5&34.5&\textbf{47.2}& knife &0.9&1.4&\textbf{12.0}\\
motorcycle &59.1&\textbf{66.2}&63.7& spoon &0.0&1.4&\textbf{6.6} \\
airplane &26.9&60.3&\textbf{64.7}& bowl &7.6&19.5&\textbf{22.4}\\
bus &52.4&63.1&\textbf{64.5}& banana&52.0&46.9&\textbf{63.2} \\
train&42.4&57.3&\textbf{64.5}& apple &28.8&40.4&\textbf{44.4} \\
truck &36.9&38.9&\textbf{44.8}& sandwich&37.4&39.4&\textbf{39.7} \\
boat &23.5&30.1&\textbf{42.3}& orange &52.0&52.9&\textbf{63.0} \\
traffic light &13.3&40.4&\textbf{49.9}& broccoli &33.7&36.0&\textbf{51.2} \\
fire hydrant &45.1&72.7&\textbf{73.2}& carrot &29.0&13.9&\textbf{40.0} \\
stop sign &43.4&40.3&\textbf{76.6}& hot dog &38.8&46.1&\textbf{53.0} \\
parking meter &33.5&59.8&\textbf{64.4}& pizza &\textbf{69.8}&62.0&62.2 \\
bench &26.3&16.0&\textbf{32.8}& donut &50.8& 43.9&\textbf{55.7} \\
bird &29.9&61.0&\textbf{62.6}& cake &37.3&30.6&\textbf{47.9}\\
cat &62.1&68.6&\textbf{78.2}& chair &10.7&11.4&\textbf{22.8} \\
dog &57.5&66.9&\textbf{68.2}& couch &9.4& 14.5&\textbf{35.0} \\
horse &40.7&55.6&\textbf{65.8}& potted plant &\textbf{21.8}&2.1&13.5 \\
sheep &54.0&61.4&\textbf{70.1}& bed &34.6&20.5&\textbf{48.6} \\
cow &47.2&60.7&\textbf{68.3}& dining table &1.1&9.5&\textbf{12.9} \\
elephant &64.3&76.1&\textbf{81.6}& toilet &43.8&57.8&\textbf{63.1}\\
bear &58.9&73.0&\textbf{80.1}& tv &11.5&36.0&\textbf{47.9} \\
zebra&60.7&80.8&\textbf{83.0}& laptop &37.0&35.2&\textbf{49.5}\\
giraffe &45.1&71.6&\textbf{76.9}& mouse &0.0&\textbf{13.4}&\textbf{13.4}\\
backpack &0.0&11.3&\textbf{14.6}& remote&37.2&23.6&\textbf{41.9}\\
umbrella &46.1&35.0&\textbf{61.7}& keyboard &19.0&17.9&\textbf{49.8}\\
handbag &0.0&2.2&\textbf{4.5}& cellphone &38.1&49.9&\textbf{54.1} \\
tie &15.5&14.7&\textbf{25.2}& microwave &\textbf{43.4}&28.7&38.0\\
suitcase &43.6&31.7&\textbf{46.8}& oven &29.2&13.3&\textbf{29.9}\\
frisbee &23.2&1.0&\textbf{43.8}& toaster &0.0&0.0 &0.0\\
skis &6.5&8.1&\textbf{12.8}& sink &\textbf{28.5}&21.0&28.0 \\
snowboard &10.9&7.6&\textbf{31.4}& refrigerator &23.8&16.6&\textbf{40.1} \\
sports ball &0.6&\textbf{28.8}&9.2& book &26.3&8.7&\textbf{32.2}\\
kite &14.0&\textbf{27.3}&26.3& clock &13.4&34.4&\textbf{43.2}\\
baseball bat &0.0&\textbf{2.2}&0.9& vase &\textbf{27.1}&25.9&22.6 \\
baseball globe &0.0&\textbf{1.3}&0.7& scissors &\textbf{37.0}&16.6&32.9\\
skateboard &7.6&\textbf{15.2}&7.8& teddy bear &58.9&47.3&\textbf{61.9}\\
surfboard &17.6&17.8&\textbf{46.5}& hair drier &0.0&0.0&0.0 \\
tennis racket &38.1&\textbf{47.1}&1.4& toothbrush &11.1&1.4&\textbf{12.2}\\
\cline{5-8} 
bottle &28.4&\textbf{33.2}&31.1& \textbf{mIoU} &29.9&33.9&\textbf{42.0} \\
\bottomrule 
\end{tabular}}
\end{table*}

%% file: supp/fig/voc_seg.tex
\begin{figure*}
\begin{center}
\includegraphics[width=.9\textwidth]{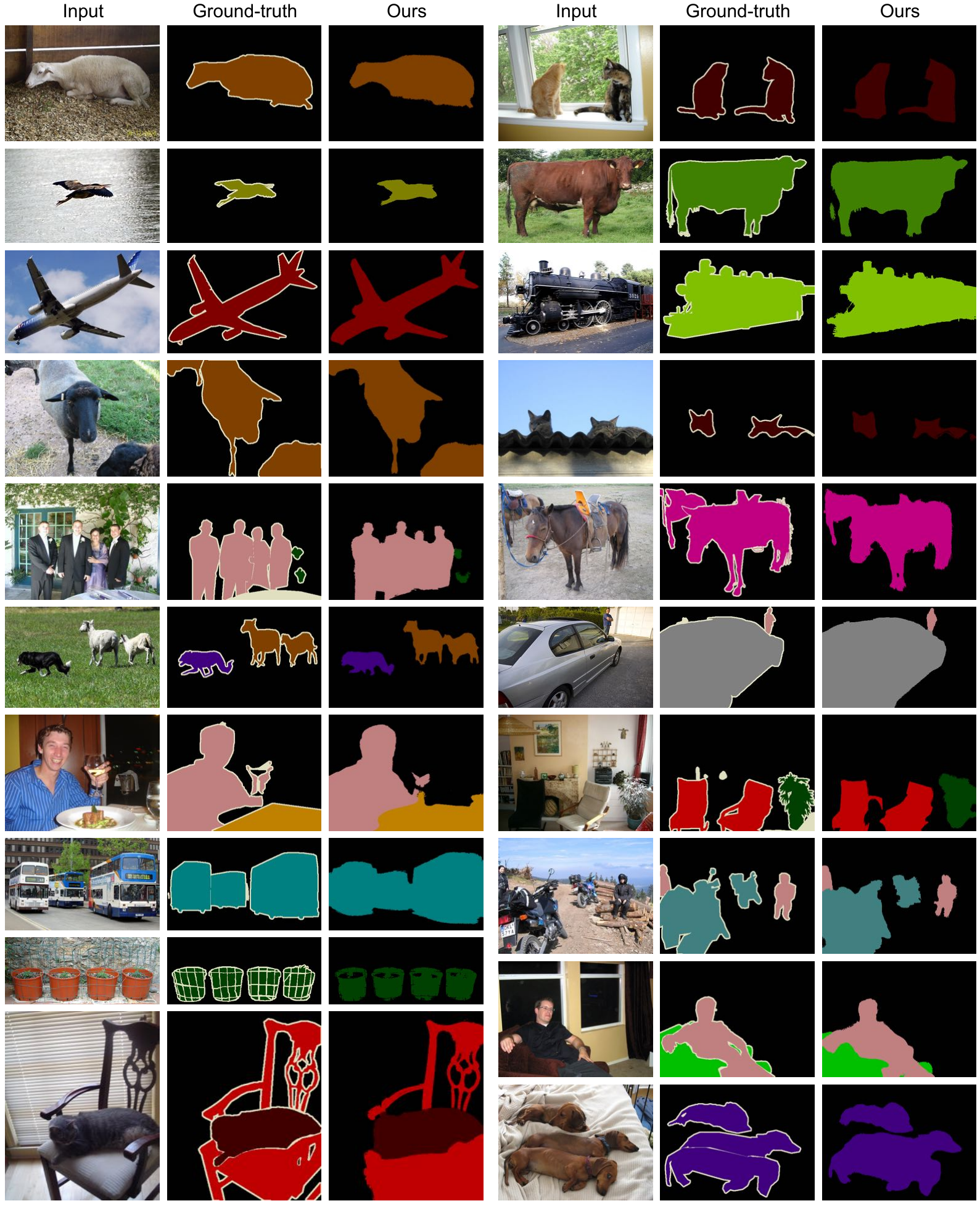}
\vspace{-18pt}
\end{center}
\caption{Qualitative segmentation results on the PASCAL VOC \textit{val} set.}
\label{fig:supp_voc_seg}
\vspace{-12pt}
\end{figure*}

%% file: supp/fig/coco_seg.tex
\begin{figure*}
\begin{center}
\includegraphics[width=.85\textwidth]{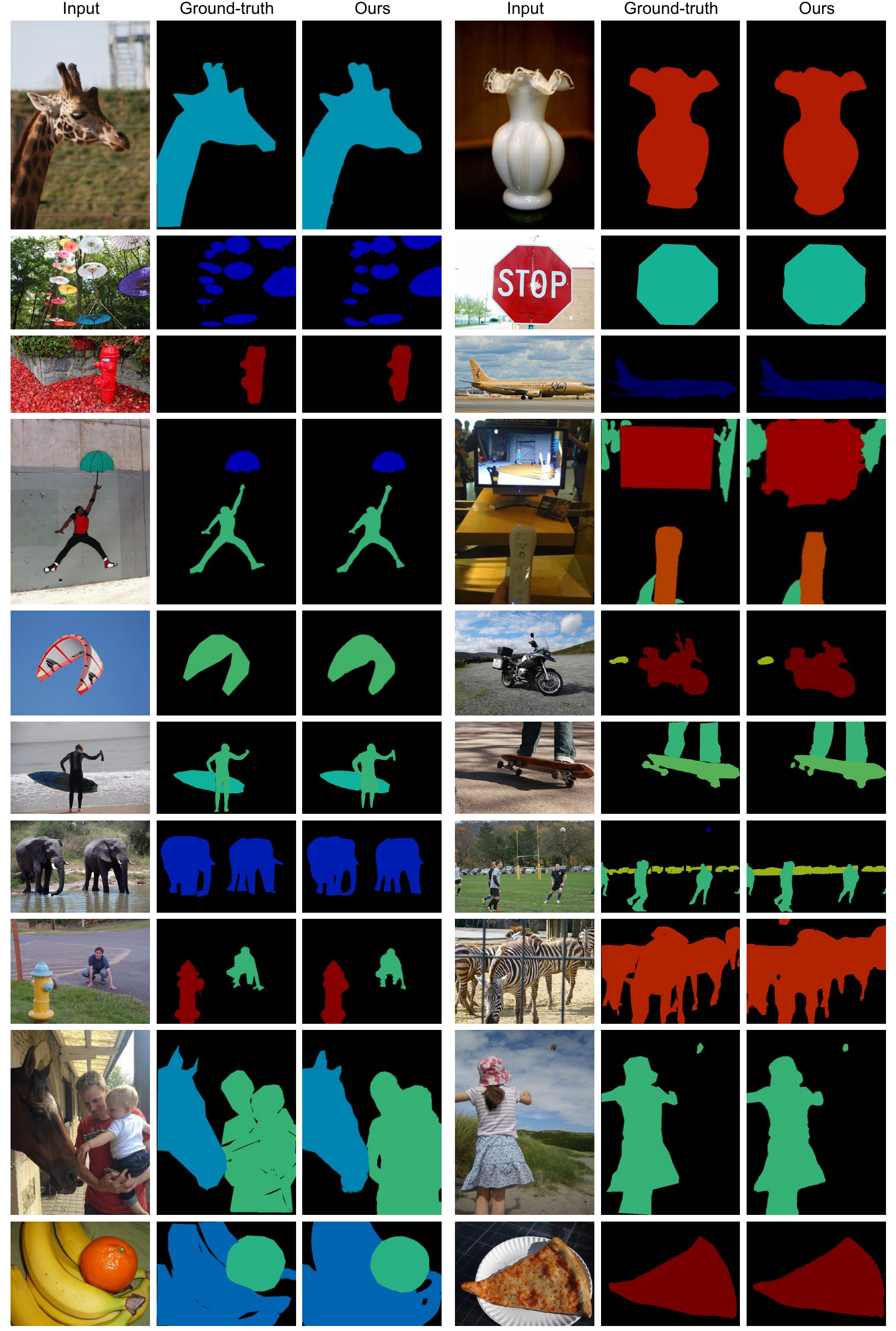}
\vspace{-18pt}
\end{center}
\caption{Qualitative segmentation results on the MS COCO \textit{val} set.}
\label{fig:supp_coco_seg}
\vspace{-12pt}
\end{figure*}

%% file: supp/fig/supp-patch-aff.tex
\begin{figure*}
\begin{center}
\includegraphics[width=0.75\textwidth]{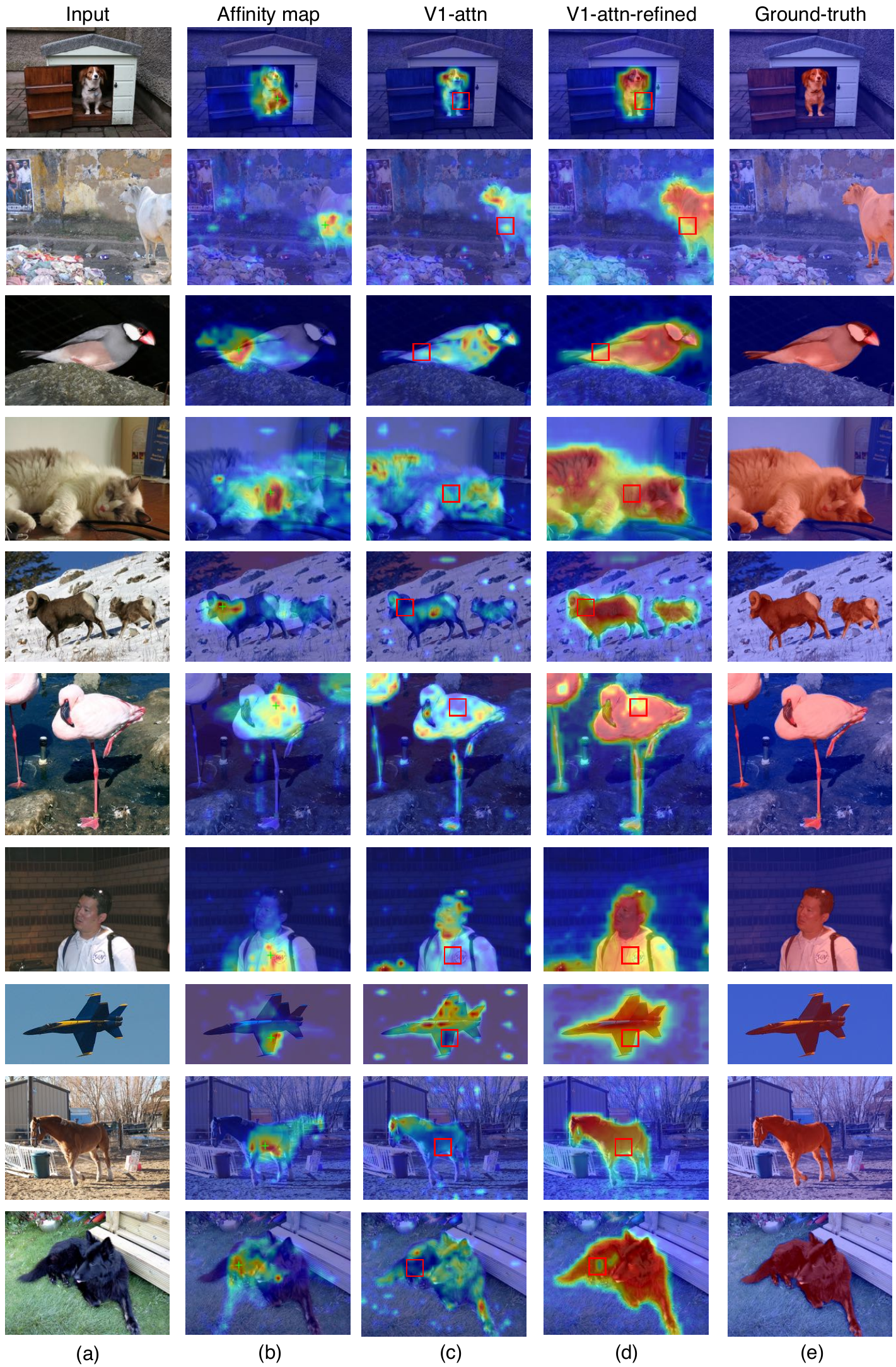}
\vspace{-18pt}
\end{center}
\caption{Visualization of the generated patch-level pairwise affinity from the proposed \nn-V1 on the PASCAL VOC \textit{train} set. (a) Input; (b) Affinity map (the generated affinity maps for the points marked by the green crosses);  (c) V1-attn (the generated transformer attention maps from \nn-V1, where the red squares denote the original attention scores for the corresponding points in (b)); (d) V1-attn-refined (the refined class-specific transformer attention maps, where the red squares denote the refined attention scores using the corresponding affinity maps of (b)); (e) Ground-truth. }
\label{fig:supp_patch_aff}
\vspace{-12pt}
\end{figure*}

%% file: supp/fig/supp-v1-cam.tex
\begin{figure*}
\begin{center}
\includegraphics[width=\textwidth]{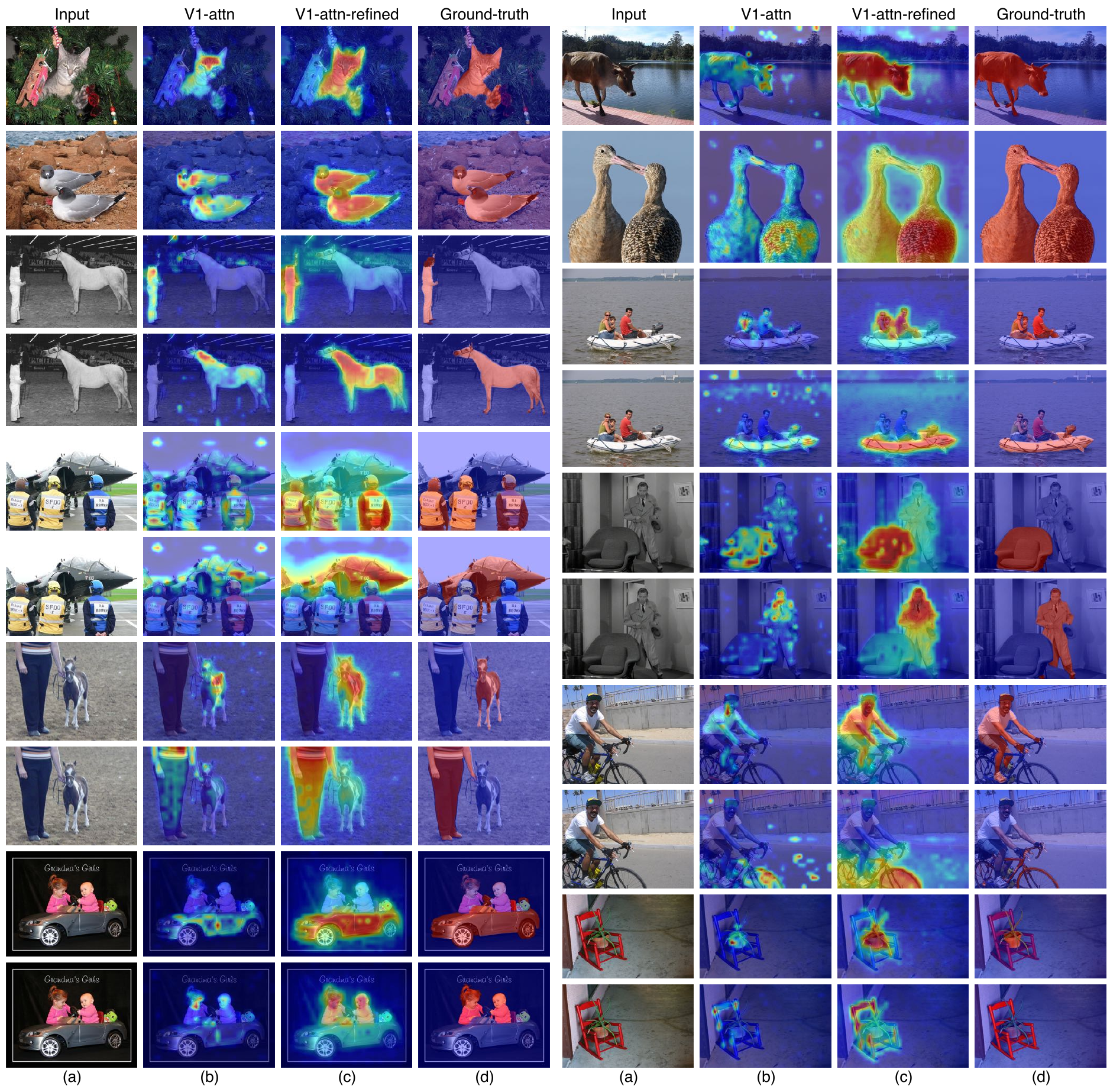}
\vspace{-18pt}
\end{center}
\caption{Visualization of the generated class-specific transformer attention maps and their refined results using the patch-level pairwise affinity from the proposed \nn-V1 on the PASCAL VOC \textit{train} set. (a) Input; (b) V1-attn (the generated class-specific transformer attention maps from the proposed \nn-V1);  (c) V1-attn-refined (the refined class-specific transformer attention maps using the patch-level pairwise affinity from the proposed \nn-V1); (d) Ground-truth. }
\label{fig:supp_v1_cam}
\vspace{-12pt}
\end{figure*}

%% file: supp/fig/supp-v2-cam-voc.tex
\begin{figure*}
\begin{center}
\includegraphics[width=0.8\textwidth]{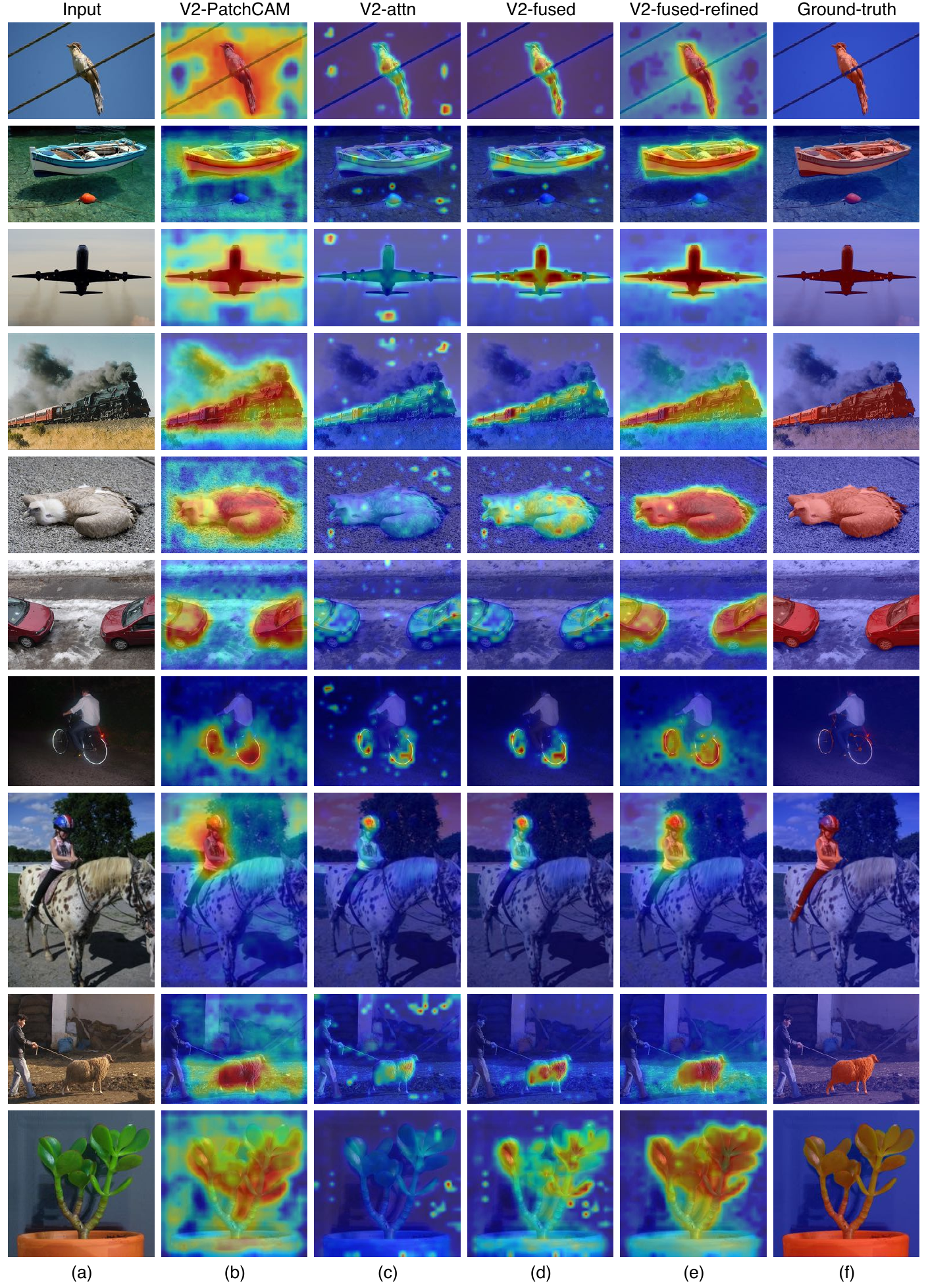}
\vspace{-18pt}
\end{center}
\caption{Visualization of the generated class-specific object localization maps from \nn-V2 on the PASCAL VOC \textit{train} set. (a) Input; (b) V2-PatchCAM (the generated PatchCAM maps from \nn-V2);  (c) V2-attn (the generated class-specific transformer attention maps from \nn-V2); (d) the fusion maps of (b) and (c); (e) the refined fusion maps by the patch-level pairwise affinity from \nn-V2; (f) Ground-truth. }
\label{fig:supp_v2_cam}
\vspace{-12pt}
\end{figure*}

%% file: supp/fig/supp-v2-cam-coco.tex
\begin{figure*}
\begin{center}
\includegraphics[width=0.78\textwidth]{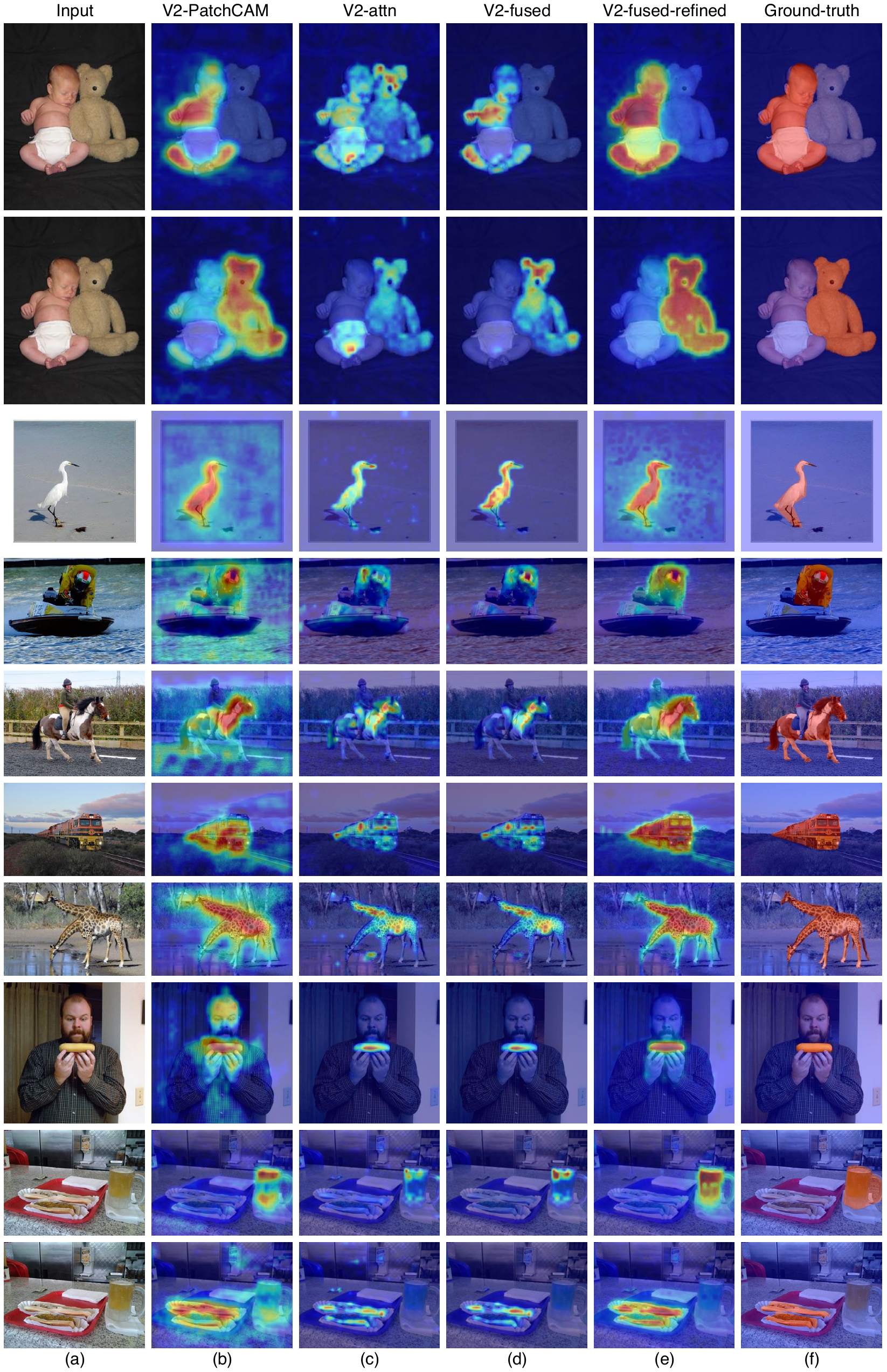}
\vspace{-18pt}
\end{center}
\caption{Visualization of the generated class-specific object localization maps from \nn-V2 on the MSCOCO \textit{train} set. (a) Input; (b) V2-PatchCAM (the generated PatchCAM maps from \nn-V2);  (c) V2-attn (the generated class-specific transformer attention maps from \nn-V2); (d) the fusion maps of (b) and (c); (e) the refined fusion maps by the patch-level pairwise affinity from \nn-V2; (f) Ground-truth. }
\label{fig:supp_v2_cam_coco}
\vspace{-12pt}
\end{figure*}